\documentclass{article} %
\usepackage{iclr2026_conference,times}
\usepackage{xspace}
\usepackage{array}

\usepackage{amsmath,amsfonts,bm}

\def\eqref#1{equation~\ref{#1}}

\def\1{\bm{1}}

\DeclareMathAlphabet{\mathsfit}{\encodingdefault}{\sfdefault}{m}{sl}
\SetMathAlphabet{\mathsfit}{bold}{\encodingdefault}{\sfdefault}{bx}{n}

\newcommand{\methodname}{\texttt{MotionStream}\xspace}

\usepackage{hyperref}
\usepackage{url}
\usepackage{graphicx}
\usepackage{amsmath, amssymb}
\usepackage{multirow}
\usepackage{booktabs}
\usepackage{caption}
\usepackage{wrapfig}

\title{MotionStream: Real-Time Video Generation\\ with Interactive Motion Controls}

\author{
\hspace{-.8mm}\bf Joonghyuk Shin$^{1,2}$\quad
\bf Zhengqi Li$^{2}$\quad
\bf Richard Zhang$^{2}$\quad
\bf Jun-Yan Zhu$^{3}$ \\[4pt]
\bf Jaesik Park$^{1}$\quad\quad
\bf Eli Shechtman$^{2}$\quad\quad
\bf Xun Huang$^{2, 4}$ \\[4pt] 
\normalfont $^1$Seoul National University, $^2$Adobe Research, $^3$Carnegie Mellon University, $^4$Morpheus AI
}

\iclrfinalcopy %
\begin{document}
\maketitle

\begin{figure}[h]
    \centering
    \includegraphics[width=\textwidth]{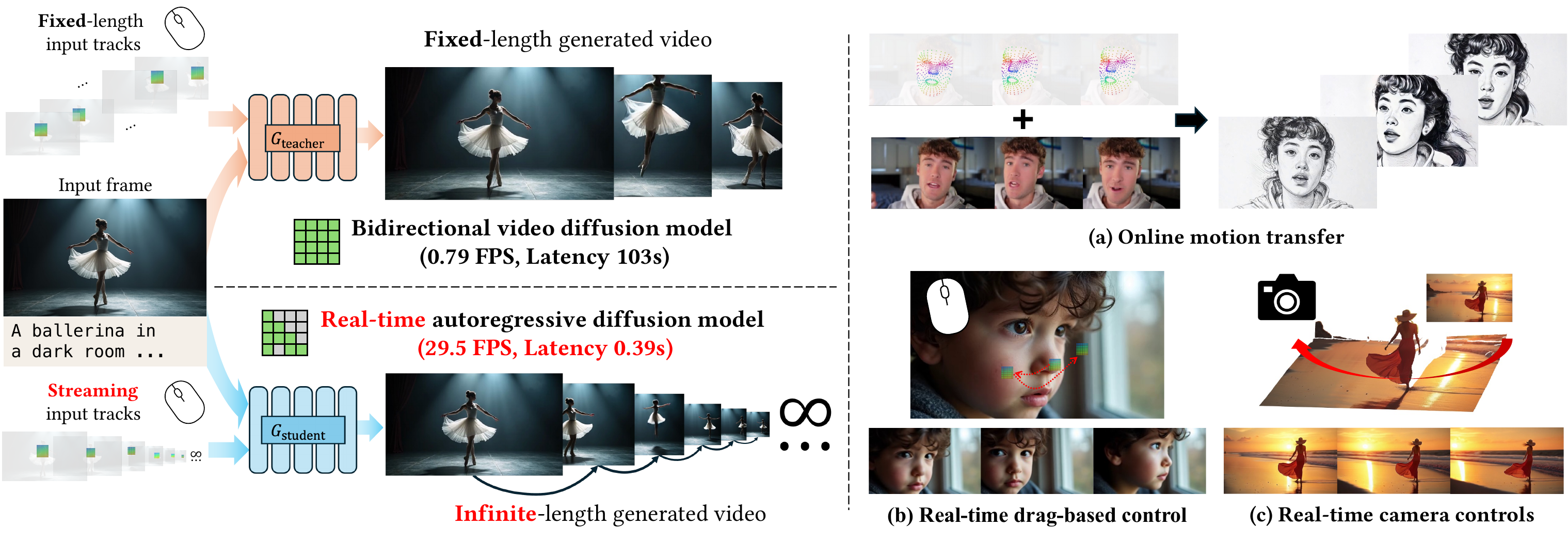}
    \caption{Prior motion-controlled video diffusion models typically operate offline to generate fixed-length sequences in parallel (top left). In contrast, our \methodname enables streaming long-video generation from a single image with track control at interactive speed (bottom left). \methodname can be applied to a variety of online downstream applications, such as real-time motion transfer, user drag operations, and 3D camera control (right).}    
    \label{fig:teaser}
\end{figure}
\begin{abstract}
Current motion-conditioned video generation methods suffer from prohibitive latency (minutes per video) and non-causal processing that prevents real-time interaction. We present \methodname, enabling sub-second latency with up to 29 FPS streaming generation on a single GPU. Our approach begins by augmenting a text-to-video model with motion control, which generates high-quality videos that adhere to the global text prompt and local motion guidance, but does not perform inference on the fly. As such, we distill this bidirectional teacher into a causal student through Self Forcing with Distribution Matching Distillation, enabling real-time streaming inference. Several key challenges arise when generating videos of long, potentially infinite time-horizons -- (1) bridging the domain gap from training on finite length and extrapolating to infinite horizons, (2) sustaining high quality by preventing error accumulation, and (3) maintaining fast inference, without incurring growth in computational cost due to increasing context windows. A key to our approach is introducing carefully designed sliding-window causal attention, combined with attention sinks. By incorporating self-rollout with attention sinks and KV cache rolling during training, we properly simulate inference-time extrapolations with a fixed context window, enabling constant-speed generation of arbitrarily long videos. Our models achieve state-of-the-art results in motion following and video quality while being two orders of magnitude faster, uniquely enabling infinite-length streaming. With \methodname, users can paint trajectories, control cameras, or transfer motion, and see results unfold in real-time, delivering a truly interactive experience.
\end{abstract}
\section{Introduction}
The ultimate goal of motion-controlled video synthesis is to give creators the power of a director's chair, allowing them to intuitively guide digital actors, objects, and cameras in real time. Although recent video diffusion models have made impressive strides toward this goal~\citep{wang2024motionctrl,geng2025motion,burgert2025go,zhang2025tora,li2025magicmotion,li2025puppet,niu2024mofa,shi2024motion,zhou2025trackgo,lei2025animateanything}, generating high-fidelity videos following user-specified motion trajectories, the experience remains far from interactive.

The promise of interactive control is currently hindered by several fundamental constraints. First, generation is too slow for interaction. For example, synthesizing a 5-second video clip with Motion Prompting~\citep{geng2025motion} takes 12 minutes, trapping users in frustrating ``render-and-wait'' cycles. Second, the process is inherently non-causal, since diffusion models process the entire sequence in parallel with bidirectional attention. A user cannot see any partial results until the entire motion specification is complete. Finally, the inability to generate more than a few seconds of video severely limits the scope for any meaningful or extended creative expression. Together, these constraints (slow, non-causal, and short-duration generation) undermine the potential for a truly interactive creative experience.

To overcome these challenges, we introduce \methodname, a method designed specifically for an interactive creative experience. Unlike conventional diffusion models that operate on the entire video sequence in parallel, \methodname is an autoregressive model that synthesizes video in a streaming manner, reacting to user-drawn motion trajectories on-the-fly.

Our approach starts with a motion-controlled teacher model that uses lightweight sinusoidal embeddings with channel-wise concatenation for trajectory conditioning, avoiding the computational overhead of ControlNet-style~\citep{zhang2023adding} architectures. Trained on both text and motion conditions, we introduce joint text-motion guidance that balances precise trajectory adherence with natural secondary motions enabled by text prompts. We then distill this teacher into a causal student through Self Forcing-style self-rollout~\citep{huang2025self}. While effective for short sequences, standard approaches drift during extended generation. Our analysis of attention patterns reveals persistent focus on initial frames alongside local temporal dependencies, similar to StreamingLLM~\citep{xiao2023efficient}. This insight drives our attention sinking mechanism with rolling KV caches, which we incorporate directly into training to properly simulate inference-time extrapolation distributions, ensuring stable, indefinite-length generation at constant latency through fixed context windows.

\methodname achieves 17 FPS at 480P and 10 FPS at 720P resolutions with sub-second latency on a single H100 GPU, reaching 29 FPS when optimized with efficient VAE decoders that we specifically train for streaming applications. Through extensive experiments and ablations, we demonstrate state-of-the-art performance across diverse motion control tasks including camera control, where our approach outperforms recent 3D methods while being more than 20$\times$ faster. \methodname transforms video generation from a passive waiting experience into an active creative process, where users can continuously interact with and guide the generation in real-time. 

Our key contributions are:
\begin{enumerate}
    \item We present the first \textit{streaming} motion-conditioned video generation pipeline capable of running at 29.5 FPS on a single H100 GPU, enabling real-time interactive applications.
    \item We propose a synergistic system harmonizing efficient architectural designs, including a lightweight track head and conditioning modules, with a distillation process that integrates joint text-motion guidance into the training objective, further accelerated by a Tiny VAE.
    \item We introduce a distillation strategy for long video generation that systematically explores attention sinks and local attention with extrapolation-aware training for the first time, effectively preventing drift during long-term streaming.
    \item Our approach achieves state-of-the-art results on motion transfer and camera control at orders of magnitude faster speeds, robustly generalizing to diverse interactive use cases.
\end{enumerate}

\section{Related Work}
\noindent\textbf{Controllable Video Generation.}
Enabling precise user control is essential for applying video generative models to diverse downstream applications~\citep{li2025diffueraser, tu2025videoanydoor, bahmani20244d, gao2024vista, wu2025video, fu2025learning}. To this end, a large body of recent research has explored various types of control signals for video generation, such as structure control~\citep{xing2024tooncrafter, yang2025layeranimate, jiang2025vidsketch, pang2024dreamdance, xing2025motioncanvas}, camera control~\citep{gao2024cat3d, zheng2024cami2v, he2024cameractrl, bai2025recammaster, wu2025cat4d, yu2025trajectorycrafter, bahmani2024vd3d, yang2024direct, zheng2024cami2v, zheng2025vidcraft3}, subject control~\citep{huang2025videomage, liu2025animateanywhere, fei2025skyreels, liu2025phantom}, and audio control~\citep{tian2024emo, gao2025wans2vaudiodrivencinematicvideo, peng2024synctalk}.

As a unique modality that captures underlying video dynamics, motion has become a key conditioning signal for recent video diffusion models. Recent video diffusion models often condition generated videos on diverse forms of motion representations, including optical flow, 2D/3D motion trajectories, bounding boxes, and semantic segmentation~\citep{goldman2008video,niu2024mofa, li2024generative, wu2024draganything, geng2025motion, zhang2025tora, gillman2025force, shi2024motion, wu2024motionbooth, gu2025diffusion, burgert2025go, tanveer2024motionbridge}.
Despite their impressive quality, these methods are fundamentally limited to offline processing because they rely on diffusion models with full bidirectional attention, which requires the entire control signal to be known in advance. This constraint prevents their use in real-time, interactive applications.

\noindent\textbf{Autoregressive Video Models.}
Early work adopted generative adversarial networks (GANs) for autoregressive or parallel video synthesis~\citep{vondrick2016generating,brooks2022generating,villegas2017decomposing,denton2017unsupervised,tulyakov2018mocogan,liu2021infinite,li2022infinitenature}. More recently, there has been a paradigm shift towards using diffusion models trained with denoising objectives~\citep{ho2022video,blattmann2023align,yang2025cogvideox,kong2024hunyuanvideo,polyak2024movie,blattmann2023stable,villegas2023phenaki,deng2024nova,gupta2024photorealistic,wan2025}, or autoregressive (AR) models trained with next-token prediction~\citep{weissenborn2020scaling,kondratyuk2024videopoet,yan2021videogpt,wang2024loong,Bruce2024GenieGI,ren2025next}.

Another line of research integrates AR and diffusion to enable causal, high-quality video generation~\citep{ruhe2024rolling,kim2024fifo,xie2024progressive,zhang2025packing,sun2025ar,weng2024art,liu2024redefining,chen2024diffusion,guo2025long,hu2024acdit,jin2024pyramidal,gu2025long,gao2024ca2,li2025arlon,zhang2025test}. Our work is inspired by the recent paradigm that distills a slow teacher model into a fast AR student for real-time performance~\citep{yin2025causvid,huang2025self,lin2025diffusion}. However, these approaches either exhibit severe color drifts beyond the training horizon or require complex long-video finetuning, which poses challenges for controllable video generation.

\noindent\textbf{Interactive Video World Model.}
Our work also belongs to interactive video world models, which aim to simulate environments for real-time interaction. This area has recently gained significant attention, as several recent works have shown impressive real-time, user-driven interaction~\citep{ball2025genie,li2025hunyuan,team2025yan,he2025matrix,bar2025navigation,po2025long}. However, most existing approaches either require substantial compute for inference~\citep{ball2025genie,parkerholder2024genie2}, or are limited to closed-domain or synthetic environments~\citep{yu2025gamefactory,guo2025mineworld,yang2024playable}. In contrast, our work demonstrates that real-time, interactive generation for open-domain, photorealistic videos can be achieved on a single GPU.

\section{\methodname: Streaming Generation Meets Motion Controls}
\begin{figure}[t]
\centering
 \includegraphics[width=1.0\textwidth]{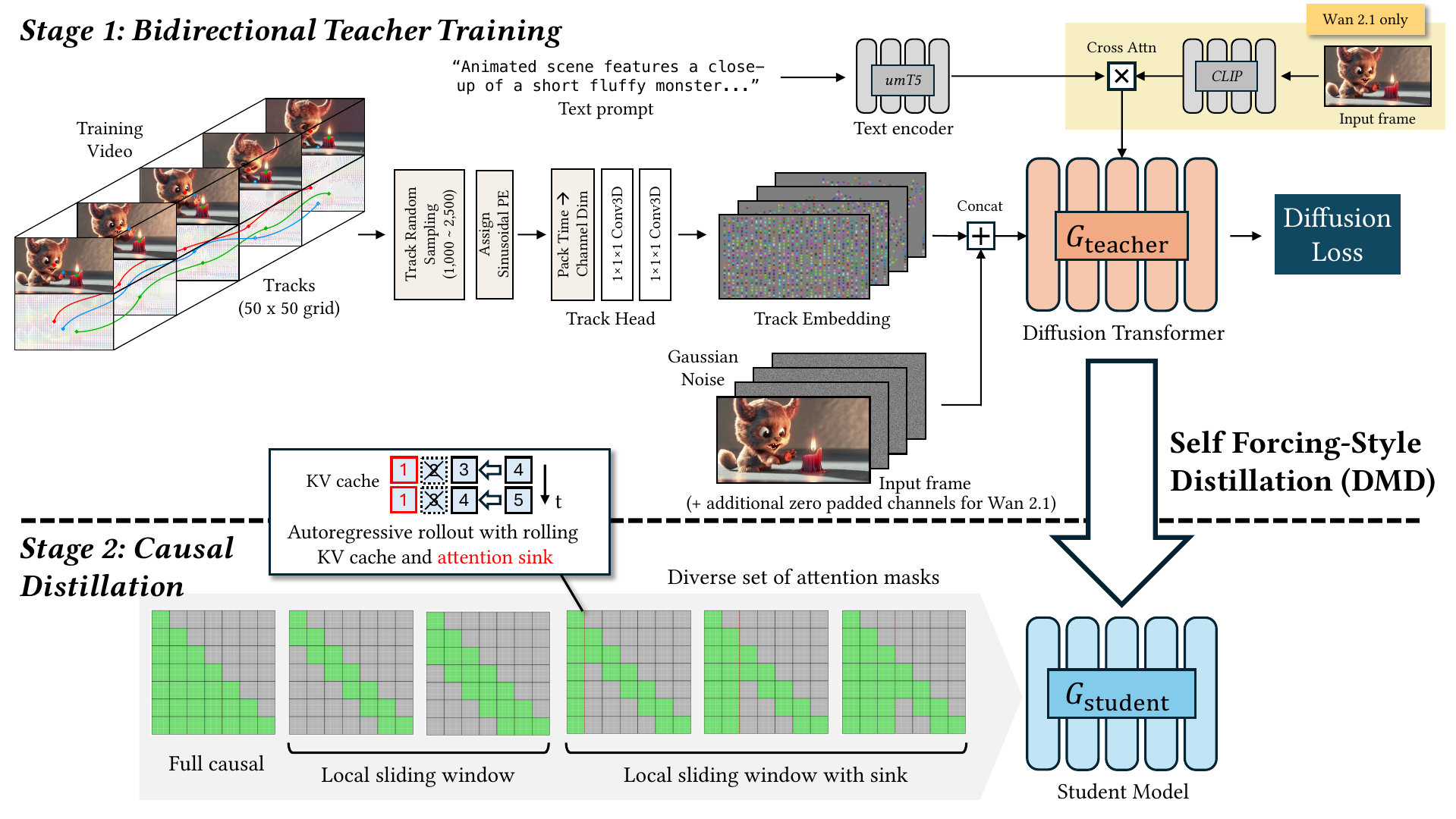}
\caption{\textbf{Model architecture and training pipeline.} 
To build a teacher motion-controlled video model, we extract and randomly sample 2D tracks from the input video and encode them using a lightweight track head. The resulting track embeddings are combined with the input image, noisy video latents, and text embeddings as input to the diffusion transformer with bidirectional attention, which is then trained with a flow matching loss (top).
We then distill a few-step causal diffusion model from the teacher through Self Forcing-style DMD distillation, integrating joint text-motion guidance into the objective, where autoregressive rollout with rolling KV cache and attention sink is applied during both training and inference (bottom).}
\label{fig:method}
\vspace{-1em}
\end{figure}

Existing motion-conditioned video generation methods achieve strong motion-video alignment, but cannot support streaming interaction since bidirectional attention requires all future control signals upfront. Our proposed \methodname addresses this through carefully designed causal distillation techniques, as illustrated in Figure~\ref{fig:method}. We first describe how to equip a pretrained video diffusion model with motion-control capability (Sec.~\ref{sec:adding_motion_control}) to serve as our bidirectional teacher, utilizing a lightweight track head and control modules designed to minimize architectural overhead. We then introduce our causal distillation pipeline, which performs extrapolation-aware training with attention sinks and local windows for long video generation, while integrating expensive joint text-motion guidance directly into the distillation objective for efficiency. Combined with our Tiny VAE, these joint efforts enable a highly responsive streaming experience.

\subsection{Adding Motion Controls to Bidirectional Teacher Models}
\label{sec:adding_motion_control}

Training a high-quality motion-conditioned teacher model is important, as it determines both the quality upper bound and the architectural efficiency of the final distilled system. The teacher must also support diverse motion modalities, from complex real-world object dynamics to camera motions and user drags, which we achieve through the following design. We build our motion-guided teacher model on top of the Wan DiT family~\citep{wan2025}.

\noindent\textbf{Track Representation and Track Head Design.} 
Following MotionPrompting~\citep{geng2025motion}, each 2D track is assigned a unique $d$-dimensional embedding vector $\phi_n$, derived from a randomly sampled ID number through sinusoidal positional encoding. While encoding tracks as an RGB video and processing it through the CausalVAE is possible, we find that representing them with sinusoidal embeddings with a learnable track head achieves superior track adherence, video quality, and faster speed. We validate this in our experiments in Table~\ref{tab:track_representation}. Given $N$ tracks $\{(x_t^n, y_t^n)\}_{n=1}^N$ across  $T$ temporal frames, the input track-conditioning signal $c_m \in \mathbb{R}^{T \times H/s \times W/s \times d}$ is constructed by placing visible track embeddings at spatially downsampled locations, where $s$ is the VAE spatial downsampling rate and $v[t,n] \in \{0,1\}$ indicates track visibility:
\begin{equation}
c_m\big[t, \lfloor \tfrac{y_t^n}{s} \rfloor, \lfloor \tfrac{x_t^n}{s} \rfloor \big] = v[t, n] \cdot \phi_n.
\end{equation}

Our lightweight track-encoding head performs $4\times$ temporal compression followed by a $1 \times 1 \times 1$ convolution. Prior methods adopt a ControlNet-style architecture~\citep{zhang2023adding, geng2025motion}, which doubles FLOPs by duplicating network blocks. Instead, we directly concatenate the processed track embeddings with video latents, requiring only minor channel adjustments in the patchifying layer while leaving the core DiT architecture unchanged. 

\noindent\textbf{Training.}
We train the motion-guided teacher model through rectified flow matching objective~\citep{Liu2022FlowSA, lipman2022flow}, where the forward process linearly interpolates between data $z_0$ and Gaussian noise $z_1 \sim \mathcal{N}(0, I)$: $z_t = (1-t) z_0 + t z_1, \  t \in [0,1]$.
The model is trained to predict the expected velocity fields with conditional flow matching loss $\mathcal{L}_{\text{FM}}$. 
One important limitation to note is that the model cannot inherently distinguish between occluded (non-visible) tracks and unspecified tracks, as both are represented by zero values. When a user releases controls during interaction, the model cannot determine whether the sudden zero values indicate occlusion or simply the absence of specification. This ambiguity occasionally leads to artifacts where objects abruptly appear or disappear. 
To address this issue, we introduce stochastic mid-frame masking with probability $p_{\text{mask}} = 0.2$, where $c_m[t_{\text{rand}}, :, :] = 0$, for randomly selected mid-frame chunks $t_{\text{rand}}$. In practice, we first train the model without masking to establish strong track-following capability, and then fine-tune with stochastic masking to preserve coherence when track signals change intermittently.

\noindent\textbf{Joint Guidance with Text and Motion Conditions.}
\label{sec:joint_guidance}
Classifier-free guidance is an effective technique for steering diffusion models.
We use both text and motion guidance and observe that they are complementary to each other. 
Text guidance generates natural dynamics but fails to maintain trajectory adherence. In contrast, track guidance enforces strict trajectory alignment but can produce overly simplistic and rigid motions, such as pure 2D planar translations in real-time drag scenarios.
Therefore, we introduce a joint combination for simultaneous text and motion guidance:
\begin{equation}
    \hat{v} = v_{\text{base}} + w_t \cdot \big( v(c_t, c_m) - v(\varnothing, c_m) \big) + w_m \cdot \big( v(c_t, c_m) - v(c_t, \varnothing) \big),
    \label{eq:joint_guidance}
\end{equation}
where $v_{\text{base}} = \alpha \cdot v(\varnothing, c_m) + (1-\alpha) \cdot v(c_t, \varnothing)$ and $\alpha = w_t/(w_t + w_m)$ (we omit $z_t$ for brevity).
We find that the joint guidance weights ($w_t = 3.0$, $w_m = 1.5$) provide a good balance: text conditioning enables realistic dynamics, even with sparse, flat-grid inputs, while track guidance preserves trajectories and maintains shape fidelity. We further analyze these effects in Sec.~\ref{sec:ablation}. Although this increases sampling cost from $2$ to $3$ function evaluations (NFE) per denoising step in the teacher model, our causal distillation (described in the next section) embeds all guidance into a single NFE, thereby eliminating this overhead in the student model.

\subsection{Causal Distillation}
\begin{wrapfigure}{r}{0.5\textwidth} %
    \vspace{-14pt}
    \centering
    \includegraphics[width=1.0\linewidth]{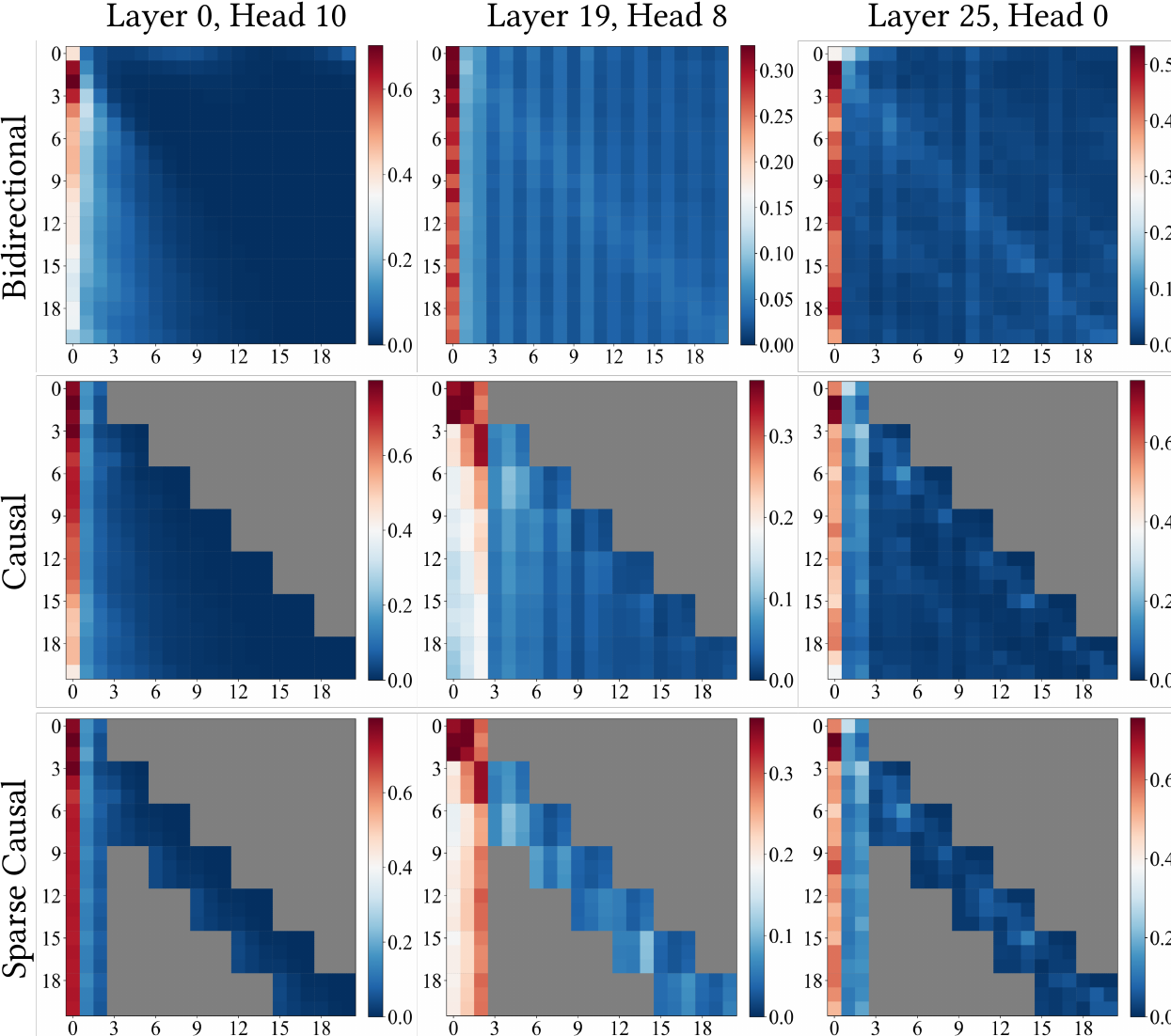}
    \caption{\textbf{Visualization of self attention probability map.} 
    We visualize attention probability maps for bidirectional, full causal, 
    and causal sliding window attentions. Several attention heads focus on 
    the tokens corresponding to the initial frame throughout denoising generation.}
    \label{fig:attention_map}
    \vspace{-12pt}
\end{wrapfigure}

\label{sec:causal_distillation}
Existing motion-controlled video diffusion models require around 50 denoising steps to generate high-quality video and also presume that the entire input motion trajectory is known before generation.
In this section, we distill the slow teacher model into a causal video diffusion model, enabling real-time streaming of long videos with motion control. Our training pipeline starts from the Self Forcing paradigm~\citep{huang2025self}, but off-the-shelf, we find this exhibits large latency fluctuations due to varying attention window sizes and performs well only within the teacher’s training horizon (\textit{i.e.}, 81 frames), with quality quickly degrading when extrapolating to longer sequences. This section presents several key technical innovations to address these issues.

\noindent\textbf{Attention sink in video diffusion model.}
Autoregressive generation with sliding-window attention, as employed in Self Forcing~\citep{huang2025self}, is prone to quality degradation and drift during long-video extrapolation. To understand this failure, we visualize the self-attention maps in Figure~\ref{fig:attention_map} for both bidirectional and causal attention. Notably, many attention heads focus on the initial tokens corresponding to the input image. This phenomenon mirrors the observations in large language models~\cite{xiao2023efficient}, where initial tokens play a crucial role in stable streaming generation.
Inspired by this, we adapt the attention sink concept to our video model by maintaining a local attention window while preserving the initial frame's tokens as a fixed anchor during both training and inference. In our chunk-wise autoregressive setup, this sinking mechanism prevents the model from drifting during long-video extrapolation, even when trained only on short sequences limited by the teacher model.

\noindent\textbf{Causal Adaptation.}
Following the initialization protocol from CausVid~\citep{yin2025causvid}, we start our student model with the weights of the motion-guided teacher diffusion model and adapt it for causal attention architecture and few-step trajectory, using regression on ODE solution pairs sampled from the teacher. We train the model using attention masks with varying context window size and attention sink size for better generalization.

\noindent\textbf{Self Forcing-Style Distillation.}
Following Self Forcing~\citep{huang2025self}, we perform temporal autoregressive roll-out with distribution matching distillation (DMD)~\citep{yin2024one, yin2024improved}. During roll-out, video chunks are generated sequentially, conditioned on previously self-generated outputs rather than ground truth using KV cache.

We first define a video latent divided into $L$ chunks as $\{z^{i}_{t}\}_{i=1}^L$, where $t$ is the timestep. The sampling process of the $i$-th chunk can attend to its own noisy tokens and previously generated clean key and value tokens: $\mathcal{C}_i = \{z_t^{i}\} \cup \{z_0^j\}_{j \leq S} \cup \{z_0^{j}\}_{\max(1, i-W) \leq j < i}$, where $S$ denotes the number of sink chunks, and $W$ the attention window. Setting it to $\{z_t^{i}\} \cup \{z_0^j\}_{j<i}$ would correspond to a causal case with full history context window, matching that of Self Forcing. The generator $G_\theta$ produces each chunk autoregressively through the factorization $p_\theta(z_0^{1:L}) = \prod_{i=1}^{L} p_\theta(z_0^i | \mathcal{C}_i)$. Sampling each $z^{i}_{0}$ involves $K$-step iterative denoising on its latent $z^{i}_{t}$, starting from pure noise $z_T^{i} \sim \mathcal{N}(0, I)$.

In our approach, because the KV cache is continuously updated, RoPE values are assigned based on cache position rather than absolute temporal index. We apply KV cache rolling with attention sinks during both training and inference, implementing local attention without explicit attention masking and fully bridging the train–test gap, even in \emph{extrapolation} scenarios.
While TalkingMachines~\citep{low2025talkingmachines} also explores attention sinks, our method differs in two key aspects that enhance long-video stability. First, we eliminate the train-inference mismatch by explicitly simulating the extrapolation process during training, using self-rollout with a rolling KV cache and attention sinks.
In contrast, TalkingMachines employs synchronized denoising with causal attention masks, which does not fully replicate the dynamics of autoregressive inference.
Second, our training process ensures the teacher model always evaluates continuous video frames. The setup in TalkingMachines introduces a temporal discontinuity between the sink frame and subsequent frames, which can push the input outside the teacher's pre-training distribution. By maintaining continuity, our teacher provides more robust scores for distillation.

After generating all $L$ chunks through self-rollout, we obtain the complete video $\hat{z}_0 = \{z_0^{1}, \ldots, z_0^{L}\}$. We then apply the DMD objective to this entire sequence, which minimizes the reverse KL divergence between the generator's output distribution and the data distribution: $\mathcal{L}_{\text{DMD}} = \mathbb{E}_t \left[ D_{\text{KL}}(p_{t}^\text{gen} \| p_{t}^{\text{data}}) \right]$. The gradient with respect to the generator parameters $\theta$ becomes:
\begin{equation}
    \nabla_\theta \mathcal{L}_{\text{DMD}} \approx -\mathbb{E}_{t, \hat{z}_0 } \left[ \left( s_{\text{real}}(\Psi(\hat{z}_0, t), t) - s_{\text{fake}}(\Psi(\hat{z}_0, t), t) \right) \cdot \frac{\partial \hat{z}_0}{\partial \theta} \right],
\end{equation} 
where $s_{\text{real}}$ is the score function for real data (approximated by the frozen bidirectional teacher) and $s_{\text{fake}}$ is the score function trained on the generator's outputs.

Intuitively, the driving gradient for the few-step causal generator $G_\theta$ comes from the difference between the estimated real and fake scores. To transfer the high-fidelity control of our joint guidance into the student without inference overhead, we define the target real score $s_{\text{real}}$ using the frozen teacher $f_\phi$ with joint guidance (omitting $z_t$ for brevity):
\begin{equation}
    s_{\text{real}} = s_{\text{base}} + w_t \cdot (f_{\phi}(c_t, c_m) - f_{\phi}(\emptyset, c_m)) + w_m \cdot (f_{\phi}(c_t, c_m) - f_{\phi}(c_t, \emptyset)),
\end{equation}
where $s_{\text{base}}$ follows the weighting defined in Eq.~\ref{eq:joint_guidance}. In contrast, we parameterize the fake score $s_{\text{fake}}$ without using any CFG through a trainable critic $f_\psi$ (which approximates the generator's score): $s_{\text{fake}} = f_\psi(c_t, c_m)$. This configuration effectively ``bakes'' the computational cost of the teacher's multi-term guidance into the distillation objective, allowing the student generator to replicate the high-quality joint-guided distribution with a single function evaluation.

We update the generator ($G_\theta$) and fake score estimator ($f_\psi$) with a ratio of 1:5~\citep{yin2024improved}, allowing the critic to better approximate generated distributions. 
To reduce the memory usage during self-rollout, we adopt Self Forcing's gradient truncation strategy: randomly sampling denoising step $k$ from $[1, K]$ and using the denoised output at k-th step as the final output, enabling gradient backpropagation only through that step. By additionally treating the cached KV from previous frames as stop-gradient context (detached), we track gradients only for the selected denoising step of the current block, effectively reducing the memory usage. 

\noindent\textbf{Inference.} 
Our inference procedure identically follows the training process to ensure a perfect train-test match. We maintain a KV cache composed of $S$ chunks from the initial frames and a fixed-size local window of recent $W$ chunks. As new tokens are generated, this local window is ``rolled'' to maintain a constant size. For positional encoding, we store the final, RoPE-applied values for the static sink tokens, while tokens in the rolling window store pre-RoPE activations and receive positional indices dynamically based on their current cache location. Because this entire mechanism is simulated during training, the model seamlessly handles the discontinuity between sink and window tokens. This approach yields two key advantages over full-context methods: (1) initial image anchoring prevents drift during long rollouts, and (2) throughput and latency remain constant regardless of generated video length. We analyze the effects of attention sink and window size in Sec.~\ref{sec:ablation} and describe our streaming pipeline in Sec.~\ref{sec:streaming_demo}.

\section{Experiments}

\subsection{Implementation Details}
We build upon image-to-video (I2V) variants of Wan 2.1 (1.3B) and Wan 2.2 (5B)~\citep{wan2025}. 
We train teacher models on OpenVid-1M~\citep{nan2024openvid} and synthetic data generated by Wan text-to-video (T2V) model: Wan 2.1 at 832×480 (70K synthetic samples) and Wan 2.2 at 1280×704 (30K synthetic samples). For causal adaptation and Self Forcing distillation, we sample input images, text prompts and 2D tracks from the synthetic datasets described above.
For computational efficiency, we track all real and synthetic videos from a 50×50 uniform grid with CoTracker3~\citep{karaev2024cotracker3}. We refer readers to 
Appendix for additional details.

\subsection{Quantitative Evaluations}
\label{sec:quantitative_eval}
\begin{table}[t]
\caption{\textbf{Benchmark on Motion Transfer (Reconstruction).}} 
\vspace{-2mm}
\centering
\scriptsize
\setlength{\tabcolsep}{3.5pt}
\label{tab:general_motion_condition}
\begin{tabular}{@{}llccccccccc@{}}
\toprule
\multirow{2}{*}{Method} & \multicolumn{1}{l}{\multirow{2}{*}{\begin{tabular}[c]{@{}c@{}}Backbone \& Resolution \end{tabular}}} & \multirow{2}{*}{FPS} & \multicolumn{4}{c}{DAVIS Validation Set} & \multicolumn{4}{c}{Sora Demo Subset} \\
\cmidrule(lr){4-7} \cmidrule(lr){8-11}
& & & PSNR & SSIM & LPIPS & EPE & PSNR & SSIM & LPIPS & EPE \\
\midrule
Image Conductor~\citep{li2025image} & AnimateDiff (256P) & 2.98 & 11.30 & 0.214 & 0.664 & 91.64 & 10.29 & 0.192 & 0.644 & 31.22 \\
Go-With-The-Flow~\citep{burgert2025go} & CogVideoX-5B (480P) & 0.60 & 15.62 & 0.392 & 0.490 & 41.99 & 14.59 & 0.410 & 0.425 & 10.27 \\
Diffusion-As-Shader~\citep{gu2025diffusion} & CogVideoX-5B (480P) & 0.29 & 15.80 & 0.372 & 0.483 & 40.23 & 14.51 & 0.382 & 0.437 & 18.76 \\
ATI~\citep{wang2025ati} & Wan 2.1-14B (480P) & 0.23 & 15.33 & 0.374 & 0.473 & 17.41 & 16.04 & 0.502 & 0.366 & 6.12 \\
\midrule
Ours Teacher (Joint CFG) & Wan 2.1-1.3B (480P) & 0.79 & \textbf{16.61} & \textbf{0.477} & \textbf{0.427} & \textbf{5.35} & \textbf{17.82} & \textbf{0.586} & \underline{0.333} & \textbf{2.71} \\
Ours Causal (Distilled) & Wan 2.1-1.3B (480P) & \textbf{16.7} & 16.20 & 0.447 & 0.443 & \underline{7.80} & 16.67 & 0.531 & 0.360 & 4.21 \\
Ours Teacher (Joint CFG) & Wan 2.2-5B (720P) & 0.74 & 16.10 & \underline{0.466} & \textbf{0.427} & 7.86 & \underline{17.18} & \underline{0.571} & \textbf{0.331} & \underline{3.16} \\
Ours Causal (Distilled) & Wan 2.2-5B (720P) & \underline{10.4} & \underline{16.30} & 0.456 & \underline{0.438} & 11.18 & 16.62 & 0.545 & 0.343 & 4.30 \\
\bottomrule
\vspace{-6.5mm}
\end{tabular}
\end{table}

\noindent\textbf{Motion Transfer.} 
We evaluate motion-following capability on two datasets: the DAVIS~\citep{Perazzi2016} validation set (30 videos) and 20 curated videos from the Sora webpage~\citep{videoworldsimulators2024}. 
We use both because DAVIS presents challenging sequences with significant occlusions, while the Sora set provides clean examples with consistent visibility, ensuring a comprehensive evaluation.
We directly compare the synthesized results with the corresponding ground-truth video frames. Visual fidelity is measured using PSNR, SSIM, and LPIPS~\citep{zhang2018unreasonable}, while motion accuracy is assessed via End-Point Error (EPE), computed as the L2 distance between visible input tracks and the tracks extracted from the generated videos. All models are evaluated in their optimal configuration, and evaluations are performed at $832 \times 480$ resolution after resizing. Speed measures are based on a single H100 GPU. We refer readers to Appendix for detailed protocols.

\noindent\textbf{Camera Control.}
To assess our model’s ability to generate videos following camera controls, we evaluate its zero-shot performance on single-image 3D novel view synthesis. We compare it against several recent diffusion- and feed-forward-based view synthesis baselines~\citep{zhou2025stable, yu2024viewcrafter, xu2025depthsplat} on the LLFF dataset~\citep{mildenhall2019local}. To adapt our 2D track-controlled model for this task, we first estimate scene geometry using a monocular depth network~\citep{wang2025moge} and compute a single scale factor to align the predicted depth with the reconstructed scene point cloud from COLMAP~\citep{schonberger2016structure}. We then derive 2D motion trajectories by interpolating between the input and target frame using depth and camera parameters. 
\begin{wraptable}{r}{0.6\textwidth}
\vspace{-3pt}
\centering
\scriptsize
\setlength{\tabcolsep}{5.0pt}
\caption{\textbf{Evaluation on Novel View Synthesis.} }
\label{tab:camera_control}
\begin{tabular}{lccccc}
\toprule
\multirow{2}{*}{Method} & \multirow{2}{*}{Resolution} & \multirow{2}{*}{FPS} & \multicolumn{3}{c}{LLFF} \\
\cmidrule(lr){4-6} & & & PSNR & SSIM & LPIPS \\
\midrule
DepthSplat~\citep{xu2025depthsplat} & 576P & 1.40 & 13.9 & 0.28 & 0.30 \\
ViewCrafter~\citep{yu2024viewcrafter} & 576P & 0.26 & 14.0 & 0.30 & 0.30 \\
SEVA~\citep{zhou2025stable} & 576P & 0.20 & 14.1 & 0.30 & 0.29 \\
\midrule
Ours Teacher (1.3B) & 480P & 0.79 & \textbf{16.0} & \textbf{0.42} & \textbf{0.21} \\
Ours Causal (1.3B) & 480P & \textbf{16.7} & \underline{15.7} & 0.38 & 0.23 \\
Ours Teacher (5B) & 720P & 0.74 & 14.0 & \underline{0.40} & \underline{0.22} \\
Ours Causal (5B) & 720P & \underline{10.4} & 15.0 & 0.39  & 0.23 \\
\bottomrule
\end{tabular} \label{table:camera}
\vspace{-5pt}
\end{wraptable}
For all methods, the input image aspect ratio is fixed at 16:9, and all synthesized frames are resized to $512 \times 288$ before evaluation with PSNR, SSIM, and LPIPS metrics. 
As shown in Table~\ref{table:camera}, our track-conditioned video generation models outperform other 3D novel view synthesis baselines by a large margin, even though they are not specifically designed for this task. Moreover, our causal models achieve significantly higher generation throughput compared with both the baselines and their bidirectional counterparts.

\subsection{Ablation Experiments}
\label{sec:ablation}
~\begin{wraptable}{r}{0.58\textwidth}
\vspace{-4mm}
\centering
\scriptsize
\setlength{\tabcolsep}{3.5pt}
\caption{\textbf{Comparing track representation methods.} Our sinusoidal PE with learnable track head outperforms RGB-VAE in both quality and efficiency, achieving 40× faster encoding critical for real-time streaming.}
\label{tab:track_representation}
\begin{tabular}{lccccc}
\toprule
\multirow{2}{*}{Method} & \multirow{2}{*}{\begin{tabular}[c]{@{}c@{}}Time\\(ms)\end{tabular}} & \multicolumn{4}{c}{DAVIS~/~Sora} \\
\cmidrule(lr){3-6}
 & & PSNR & SSIM & LPIPS & EPE \\
\midrule
RGB-VAE & 1053 & 16.03~/~16.99 & 0.433~/~0.544 & 0.463~/~0.363 & 8.57~/~3.96 \\
PE-Head & \textbf{24.8} & \textbf{16.29~/~17.15} & \textbf{0.452~/~0.559} & \textbf{0.456~/~0.359} & \textbf{6.54~/~3.13} \\
\bottomrule
\end{tabular}
\vspace{-2mm}
\end{wraptable}
\noindent\textbf{Track Representation.}
We compare our sinusoidal position encoding with a learnable track head against the RGB encoding strategy using a frozen VAE, following prior work~\citep{gu2025diffusion}, where each 2D track is assigned a unique RGB color vector and placed onto a canvas before being fed into the VAE. Table~\ref{tab:track_representation} shows that our method (PE-Head) outperforms RGB-VAE in both efficiency and quality. Specifically, our lightweight PE-based encoding achieves better motion alignment while being two orders of magnitude faster than the VAE-based approach.
We hypothesize sinusoidal encoding preserves stronger identification signals compared to RGB encoding due to richer expressive dimensions. 

\begin{figure}[t]
    \begin{minipage}[c]{0.48\textwidth}
        \centering
        \includegraphics[width=\textwidth]{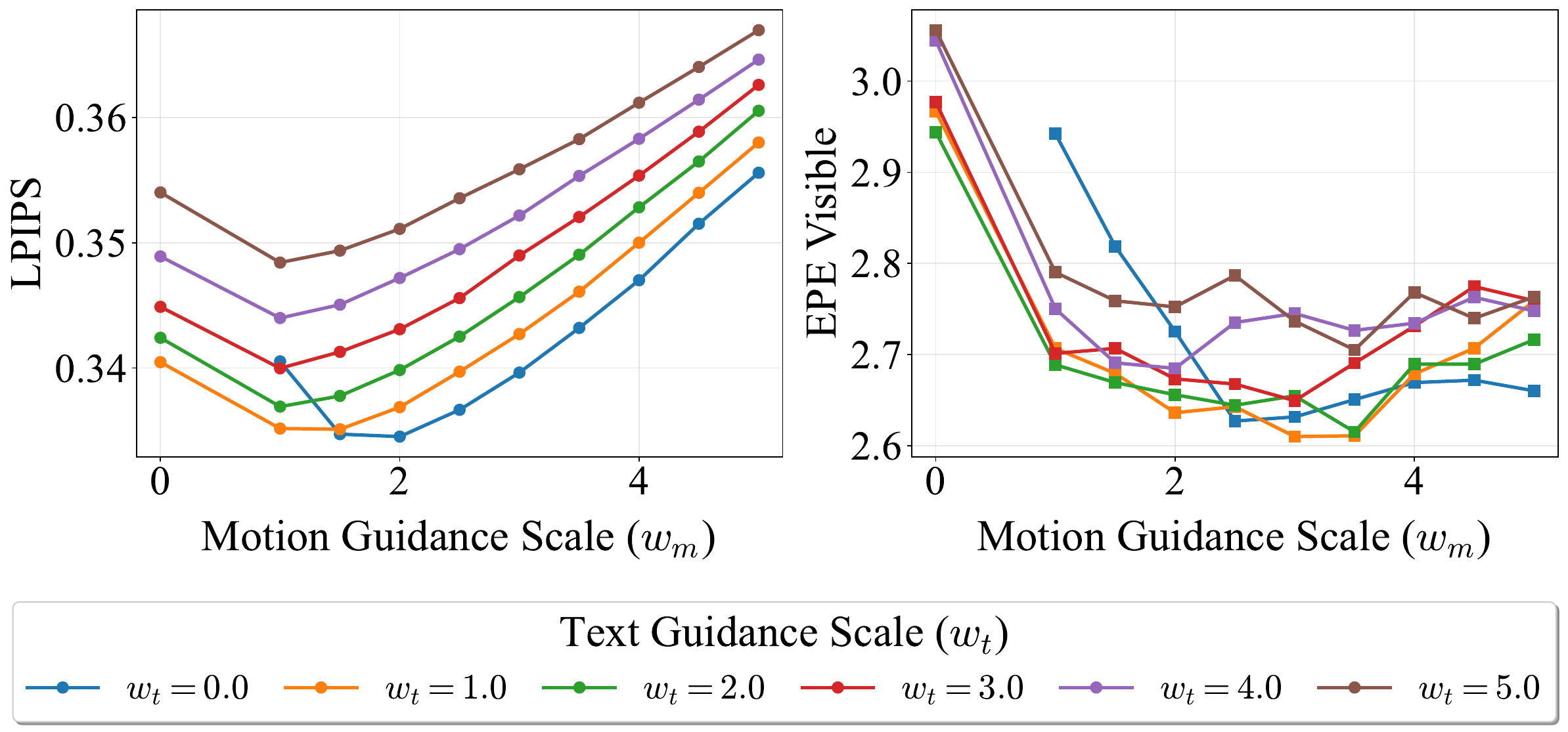}
        \caption{\textbf{Quantitative ablation on guidance.} We use Sora subset to ablate guidance strategies. Higher text guidance reduces overall metrics while motion guidance improves trajectory accuracy at the cost of visual quality (LPIPS).}
        \label{fig:guidance_ablation_quan}
        \end{minipage}
    \hfill
    \begin{minipage}[c]{0.5\textwidth}
        \centering
        \includegraphics[width=\textwidth]{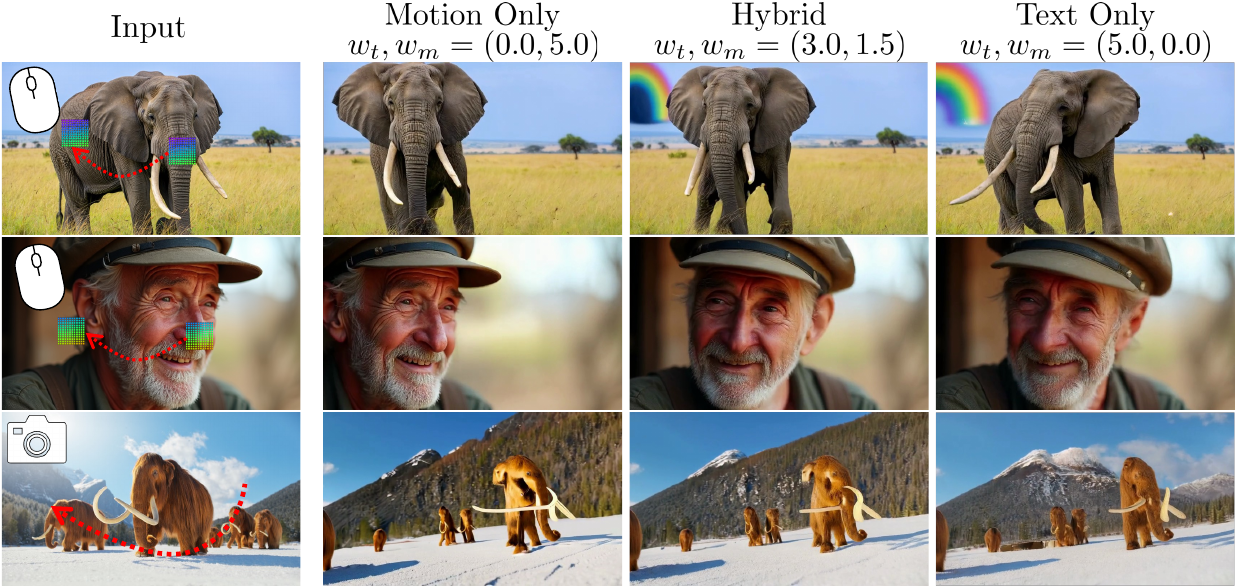}
        \caption{\textbf{Qualitative ablation on guidance.} Pure motion guidance produces rigid movements while text guidance enables natural motion and shape preservation even with imperfect tracks. Our \textit{Hybrid} joint guidance balances these two.}
        \label{fig:guidance_ablation_qual}
    \end{minipage}
\vspace{-5mm}
\end{figure}

\noindent\textbf{Guidance Strategies.}
Here we ablate our joint guidance approach from Sec.\ref{sec:joint_guidance}. While pure motion guidance ($w_t=0$, $w_m>0$) achieves the highest trajectory accuracy, as shown in Figure~\ref{fig:guidance_ablation_quan}, text guidance provides additional benefits for generating more diverse and realistic results. For example, text captions enable dynamics beyond trajectories alone, such as weather changes or object appearances as shown in the first row of Figure~\ref{fig:guidance_ablation_qual}, which illustrates dragging an elephant while prompting “rainbow appears in background”.
Our empirical setting ($w_t=3.0$, $w_m=1.5$) balances motion fidelity with natural dynamics, adapting equally well to both precise and imperfect trajectories.

\begin{figure}[t]
\centering
\includegraphics[width=1.0\textwidth]{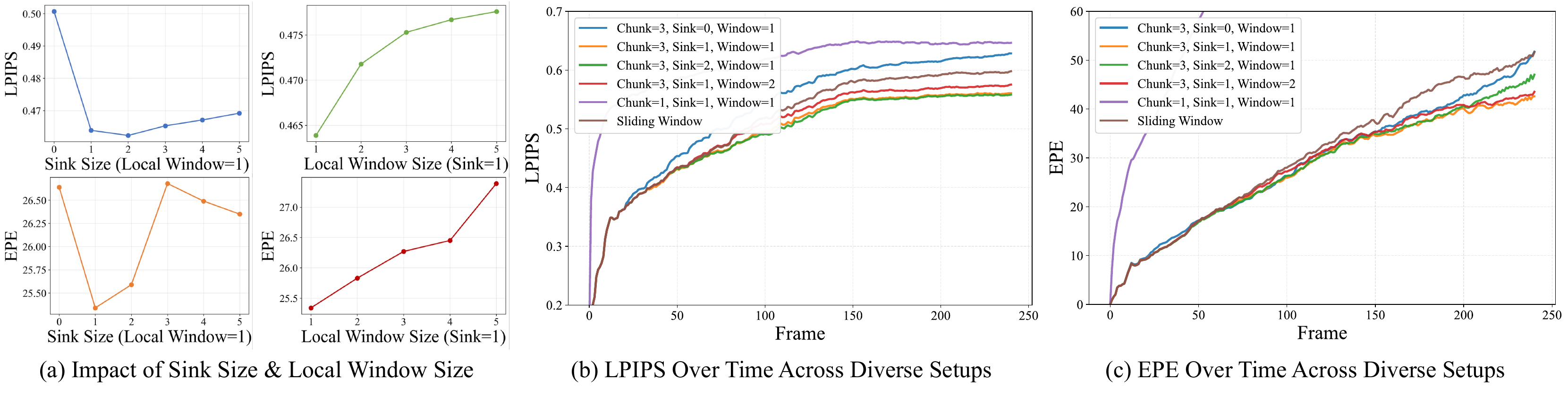}
\caption{\textbf{Impact of Sparse Attention Patterns.} Using longer clips (up to 241 frames) from the Sora subset, we ablate attention sink size and local window size in extrapolation scenarios. Having at least a single sink chunk is crucial, but more provides marginal benefit, while larger window sizes degrade performance as attending to long-past history allows errors to accumulate in context tokens.}
\label{fig:streaming_abl}
\end{figure}

\noindent\textbf{Impact of Chunk Size, Attention Sink, and Window Size.}
We investigate three key design choices that govern streaming quality and interactivity: the latent chunk size, the attention sink size, and the local context window size.

The chunk size, which defines how many video latents are processed in parallel, presents a critical trade-off between quality, latency, and throughput. As illustrated in the latency-throughput analysis in Figure~\ref{fig:sup_abl_speed_steps}, small chunks lead to lower throughput, while large chunks ($>$3) introduce prohibitive latency for real-time interaction.
Additionally, as shown in Tab.~\ref{tab:ablation}, a chunk size of 1 causes significant quality degradation.
We, therefore, select a chunk size of 3 as our optimal configuration, with further analysis provided in Appendix Sec.~\ref{sec:sup_abl}.

\begin{wraptable}{r}{0.6\textwidth}
\vspace{-4mm}
\scriptsize
\centering
\setlength{\tabcolsep}{3.5pt}
\caption{\textbf{Ablation study on Sora Extended.} \texttt{c3s1w1} maintains high visual quality with small latency \& throughput fluctuations, while vanilla sliding window exhibits large fluctuations degrading streaming stability.}
\label{tab:ablation}
\begin{tabular}{lcccc}
\toprule
\multirow{2}{*}{Config} & \multicolumn{4}{c}{Sora Extended} \\
\cmidrule(lr){2-5}
& LPIPS & EPE & Latency (s) & Throughput (FPS) \\
\midrule
Ours base (\texttt{c3s1w1}) & \textbf{0.464} & \textbf{25.34} & 0.70 $\pm$ 0.01 & \underline{16.92 $\pm$ 0.80} \\
+ Remove sink (\texttt{c3s0w1}) & 0.501 & \underline{26.64} & \underline{0.68 $\pm$ 0.005} & \textbf{17.43 $\pm$ 0.88}  \\
+ Chunk Size 1 (\texttt{c1s1w1}) & 0.597 & 76.21 & \textbf{0.30 $\pm$ 0.01} & 13.26 $\pm$ 1.36 \\
Sliding window & \underline{0.480} & 28.09 & 0.80 $\pm$ 0.08 & 14.96 $\pm$ 1.42 \\
\bottomrule
\end{tabular}
\end{wraptable}

To assess the impact of attention sinks and window size on model performance, we train a single model with randomly sampled sink and window sizes and then generate videos under different configuration combinations. We evaluate long-video extrapolation on Sora videos up to 241 frames (average 194 frames). Surprisingly, we find that the minimal configuration (a single-chunk sink with a single-chunk window) achieves the best performance. Figure~\ref{fig:streaming_abl}(a) shows that additional sinks provide only marginal gains while increasing latency, and expanded windows actually degrade performance. We hypothesize that this phenomenon arises because restricting the context to immediate predecessors, rather than long-past history, prevents error accumulation and thus reduces drift for long-video generation.

We additionally show reconstruction accuracy and throughput as well as their time evolution in Table~\ref{tab:ablation} and Figure~\ref{fig:streaming_abl}(b,c) respectively.
We denote our configuration as \texttt{chunk-3, sink-1, window-1} (abbreviated \texttt{c3s1w1}), while sliding window approach used in Self Forcing is \texttt{c3s0w6}, attending to a maximum of 6 previous chunks without sink tokens, with unbounded RoPE positions that scale with temporal frame count. Although removing sink tokens (\texttt{c3s0w1}) yields marginal speed improvements, this  comes at the cost of degraded long-term generation stability.

\begin{figure}[t]
\centering
\includegraphics[width=1.0\textwidth]{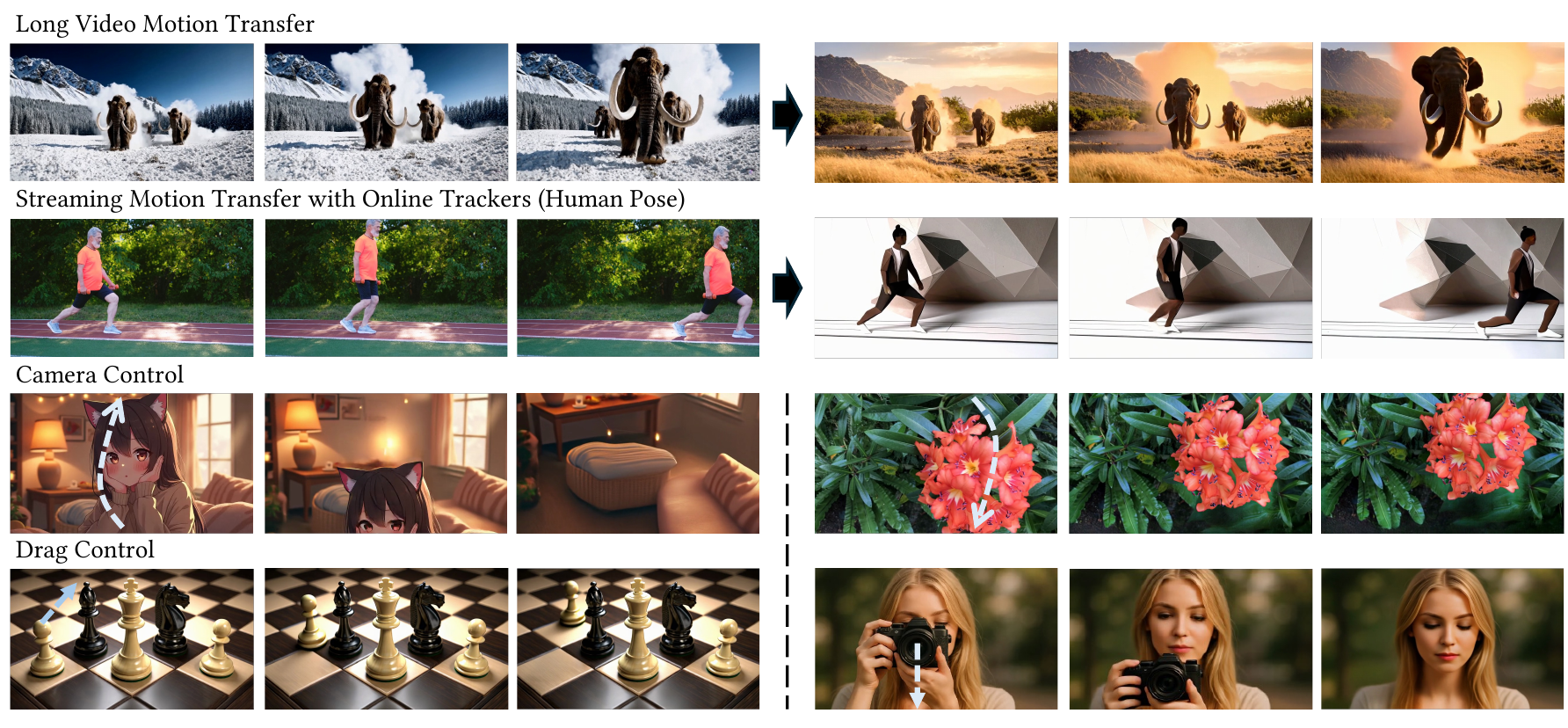}
\vspace{-6mm}
\caption{\textbf{Qualitative Results.} \methodname can perform diverse downstream applications, including long video motion transfer (from offline or online sources), drag-based controls, and precise camera control with depth estimation. We showcase a few examples here.}
\label{fig:qualitative}
\end{figure}

\begin{figure}[t]
\centering
\includegraphics[width=1.0\textwidth]{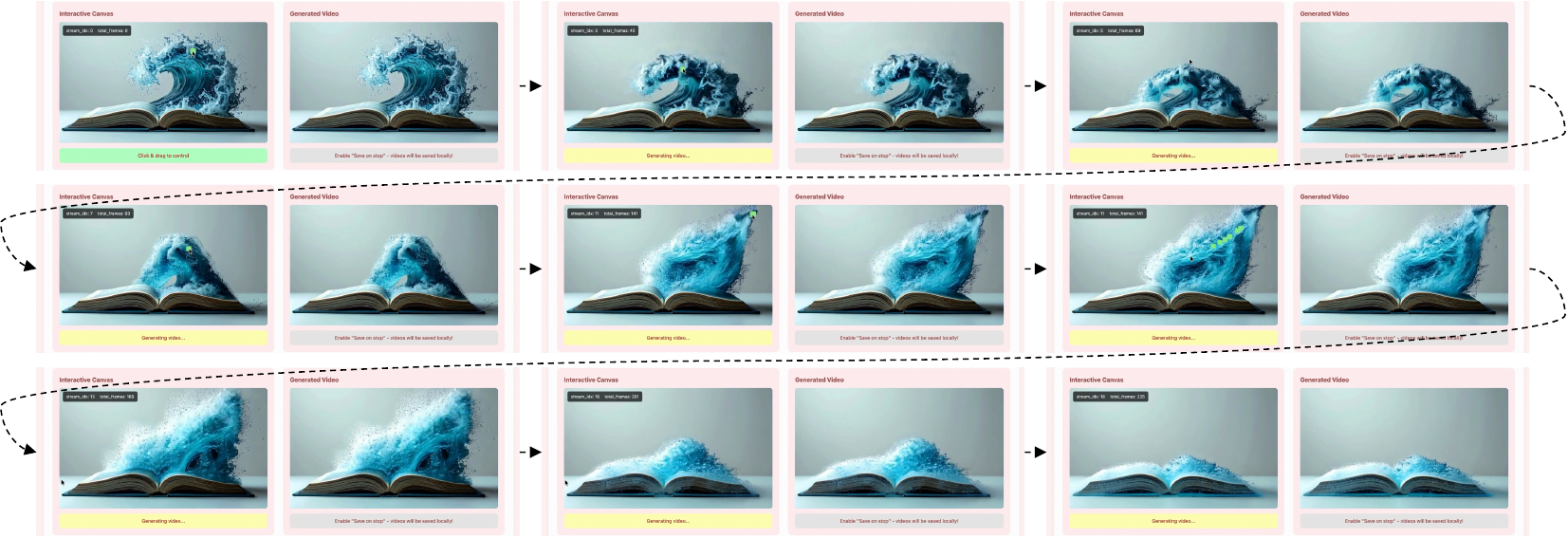}
\vspace{-4mm}
\caption{\textbf{Streaming Demo.} An example of real-time streaming generation; more details in Sec.~\ref{sec:sup_abl}.}
\label{fig:main_streaming_demo}
\vspace{-5mm}
\end{figure}

\subsection{Streaming Demo and Qualitative Results}
\label{sec:streaming_demo}
For streaming demos, we further optimize the pipeline with an efficient Tiny VAE. Inspired by~\cite{BoerBohan2025TAEHV}, we design and train a smaller VAE decoder with adversarial and LPIPS losses, regressing to the original VAE's latent space. Tiny VAE removes the VAE bottleneck by reducing the decoding time over 10$\times$, improving Wan 2.1 from 16.7 FPS with 0.69s latency to 29.5 FPS with 0.39s latency, and Wan 2.2 from 10.4 FPS with 1.1s latency to 23.9 FPS with 0.49s latency on a single H100 GPU. We provide details for the tiny VAE in Appendix~\ref{sec:sup_tiny}, noting that the performance trade-off from switching to the tiny VAE is marginal in streaming scenarios, as shown in Table~\ref{tab:vae_streaming}. 
Leveraging track representation, our method enables causal, real-time execution of capabilities typically found in motion control approaches, such as mouse-based drag control and motion transfer using tracks from online trackers. Figure~\ref{fig:qualitative} demonstrates several of these applications. Additionally, we develop an interactive demo that enables real-time user control during streaming generation. Figure~\ref{fig:main_streaming_demo} shows an example, with further details in Sec.~\ref{sec:sup_abl} and Figure~\ref{fig:sup_streaming_demo}.

\section{Conclusion}
We propose \methodname, a framework for infinite-length video generation with interactive motion control, maintaining a stable 29 FPS on a single GPU. Our contributions span from training a motion-guided teacher with efficient track head and joint text-motion guidance, to distilling it into a causal student via self-rollout with attention sink and rolling KV caches. \methodname achieves state-of-the-art results across diverse motion-conditioned generation tasks, while being significantly faster than prior methods. Limitations and future directions, as well as ethics and reproducibility statements are discussed in Appendix~\ref{sec:sup_limitation} and~\ref{sec:ethics}.

\paragraph{Acknowledgments.} This work was done at Adobe Research, during Joonghyuk Shin's internship and while Xun Huang was a Research Scientist.
Jaesik Park was supported by the IITP grant No. RS-2021-II211343 (Artificial Intelligence Graduate School Program at Seoul National University, 5\%), No. RS-2025-25442338 (AI Star Fellowship Support Program, 35\%), and the NRF grant No. RS-2024-00405857 (60\%) funded by the Korea government (MSIT). Jun-Yan Zhu was supported by the Packard Fellowship.

\clearpage
\bibliography{main}

@String(CVPR= {IEEE Conf. Comput. Vis. Pattern Recog.})

@String(ICCV= {Int. Conf. Comput. Vis.})

@String(ECCV= {Eur. Conf. Comput. Vis.})

@String(NIPS= {Adv. Neural Inform. Process. Syst.})

@String(TOG= {ACM Trans. Graph.})

@String(ICLR = {Int. Conf. Learn. Represent.})

@String(AAAI = {AAAI})

@String(CVPR  = {CVPR})

@String(ICCV  = {ICCV})

@String(ECCV  = {ECCV})

@String(NIPS  = {NeurIPS})

@String(TOG   = {ACM TOG})

@String(ICLR  = {ICLR})

@String(ICML = {ICML})

@inproceedings{wu2024draganything,
  title={Draganything: Motion control for anything using entity representation},
  author={Wu, Weijia and Li, Zhuang and Gu, Yuchao and Zhao, Rui and He, Yefei and Zhang, David Junhao and Shou, Mike Zheng and Li, Yan and Gao, Tingting and Zhang, Di},
  booktitle={European Conference on Computer Vision},
  pages={331--348},
  year={2024},
  organization={Springer}
}

@article{shah2024flashattention,
  title={Flashattention-3: Fast and accurate attention with asynchrony and low-precision},
  author={Shah, Jay and Bikshandi, Ganesh and Zhang, Ying and Thakkar, Vijay and Ramani, Pradeep and Dao, Tri},
  journal={Advances in Neural Information Processing Systems},
  volume={37},
  pages={68658--68685},
  year={2024}
}

@inproceedings{peng2024synctalk,
  title={Synctalk: The devil is in the synchronization for talking head synthesis},
  author={Peng, Ziqiao and Hu, Wentao and Shi, Yue and Zhu, Xiangyu and Zhang, Xiaomei and Zhao, Hao and He, Jun and Liu, Hongyan and Fan, Zhaoxin},
  booktitle={Proceedings of the IEEE/CVF Conference on Computer Vision and Pattern Recognition},
  pages={666--676},
  year={2024}
}

@misc{fastvideo2025, 
    title={GitHub - hao-ai-lab/FastVideo: A unified inference and post-training framework for accelerated video generation.}, 
    url={https://github.com/hao-ai-lab/FastVideo}, 
    journal={GitHub}, 
    publisher={GitHub}, 
    author={{The FastVideo Team}},
}

@inproceedings{kingma2014adam,
  title={Adam: A method for stochastic optimization},
  author={Kingma, Diederik P and Ba, Jimmy},
  booktitle=ICLR,
  year={2015}
}

@inproceedings{loshchilov2017decoupled,
  title={Decoupled weight decay regularization},
  author={Loshchilov, Ilya and Hutter, Frank},
  booktitle=ICLR,
  year={2019}
}

@inproceedings{goldman2008video,
  title={Video object annotation, navigation, and composition},
  author={Goldman, Dan B and Gonterman, Chris and Curless, Brian and Salesin, David and Seitz, Steven M},
  booktitle={Proceedings of the 21st annual ACM symposium on User interface software and technology},
  pages={3--12},
  year={2008}
}

@inproceedings{xiao2024spatialtracker,
  title={Spatialtracker: Tracking any 2d pixels in 3d space},
  author={Xiao, Yuxi and Wang, Qianqian and Zhang, Shangzhan and Xue, Nan and Peng, Sida and Shen, Yujun and Zhou, Xiaowei},
  booktitle={Proceedings of the IEEE/CVF Conference on Computer Vision and Pattern Recognition},
  pages={20406--20417},
  year={2024}
}

@misc{BoerBohan2025TAEHV,
  author = {Boer Bohan, Ollin},
  title = {TAEHV: Tiny AutoEncoder for Hunyuan Video},
  year = {2025},
  howpublished = {\url{https://github.com/madebyollin/taehv}},
}

@misc{videoxfun2024,
  author       = {{AIGC-Apps} and {Alibaba PAI Team}},
  title        = {{VideoX-Fun: A Flexible Framework for Video Generation at Any Resolution}},
  year         = {2024},
  howpublished          = {https://github.com/aigc-apps/VideoX-Fun},
}

@inproceedings{huang2024vbench,
  title={Vbench: Comprehensive benchmark suite for video generative models},
  author={Huang, Ziqi and He, Yinan and Yu, Jiashuo and Zhang, Fan and Si, Chenyang and Jiang, Yuming and Zhang, Yuanhan and Wu, Tianxing and Jin, Qingyang and Chanpaisit, Nattapol and others},
  booktitle={Proceedings of the IEEE/CVF Conference on Computer Vision and Pattern Recognition},
  pages={21807--21818},
  year={2024}
}

@article{liu2025animateanywhere,
  title={AnimateAnywhere: Rouse the Background in Human Image Animation},
  author={Liu, Xiaoyu and Yao, Mingshuai and Zhang, Yabo and Lin, Xianhui and Ren, Peiran and Li, Xiaoming and Liu, Ming and Zuo, Wangmeng},
  journal={arXiv preprint arXiv:2504.19834},
  year={2025}
}

@inproceedings{huang2025videomage,
  title={Videomage: Multi-subject and motion customization of text-to-video diffusion models},
  author={Huang, Chi-Pin and Wu, Yen-Siang and Chung, Hung-Kai and Chang, Kai-Po and Yang, Fu-En and Wang, Yu-Chiang Frank},
  booktitle={Proceedings of the Computer Vision and Pattern Recognition Conference},
  pages={17603--17612},
  year={2025}
}

@article{liu2025phantom,
  title={Phantom: Subject-consistent video generation via cross-modal alignment},
  author={Liu, Lijie and Ma, Tianxiang and Li, Bingchuan and Chen, Zhuowei and Liu, Jiawei and Li, Gen and Zhou, Siyu and He, Qian and Wu, Xinglong},
  journal={arXiv preprint arXiv:2502.11079},
  year={2025}
}

@misc{gao2025wans2vaudiodrivencinematicvideo,
   title={Wan-S2V: Audio-Driven Cinematic Video Generation},
   author={Xin Gao and Li Hu and Siqi Hu and Mingyang Huang and Chaonan Ji and Dechao Meng and Jinwei Qi and Penchong Qiao and Zhen Shen and Yafei Song and Ke Sun and Linrui Tian and Guangyuan Wang and Qi Wang and Zhongjian Wang and Jiayu Xiao and Sheng Xu and Bang Zhang and Peng Zhang and Xindi Zhang and Zhe Zhang and Jingren Zhou and Lian Zhuo},
   year={2025},
   eprint={2508.18621},
   archivePrefix={arXiv},
   primaryClass={cs.CV},
   url={https://arxiv.org/abs/2508.18621}
}

@article{fei2025skyreels,
  title={Skyreels-a2: Compose anything in video diffusion transformers},
  author={Fei, Zhengcong and Li, Debang and Qiu, Di and Wang, Jiahua and Dou, Yikun and Wang, Rui and Xu, Jingtao and Fan, Mingyuan and Chen, Guibin and Li, Yang and others},
  journal={arXiv preprint arXiv:2504.02436},
  year={2025}
}

@inproceedings{tian2024emo,
  title={Emo: Emote portrait alive generating expressive portrait videos with audio2video diffusion model under weak conditions},
  author={Tian, Linrui and Wang, Qi and Zhang, Bang and Bo, Liefeng},
  booktitle={European Conference on Computer Vision},
  pages={244--260},
  year={2024},
  organization={Springer}
}

@inproceedings{yu2025trajectorycrafter,
    author    = {Yu, Mark and Hu, Wenbo and Xing, Jinbo and Shan, Ying},
    title     = {TrajectoryCrafter: Redirecting Camera Trajectory for Monocular Videos via Diffusion Models},
    booktitle = {Proceedings of the IEEE/CVF International Conference on Computer Vision (ICCV)},
    month     = {October},
    year      = {2025},
    pages     = {100-111}
}

@inproceedings{xing2025motioncanvas,
  title={Motioncanvas: Cinematic shot design with controllable image-to-video generation},
  author={Xing, Jinbo and Mai, Long and Ham, Cusuh and Huang, Jiahui and Mahapatra, Aniruddha and Fu, Chi-Wing and Wong, Tien-Tsin and Liu, Feng},
  booktitle={Proceedings of the Special Interest Group on Computer Graphics and Interactive Techniques Conference Conference Papers},
  pages={1--11},
  year={2025}
}

@inproceedings{gu2025diffusion,
  title={Diffusion as shader: 3d-aware video diffusion for versatile video generation control},
  author={Gu, Zekai and Yan, Rui and Lu, Jiahao and Li, Peng and Dou, Zhiyang and Si, Chenyang and Dong, Zhen and Liu, Qifeng and Lin, Cheng and Liu, Ziwei and others},
  booktitle={Proceedings of the Special Interest Group on Computer Graphics and Interactive Techniques Conference Conference Papers},
  pages={1--12},
  year={2025}
}

@article{tanveer2024motionbridge,
  title={Motionbridge: Dynamic video inbetweening with flexible controls},
  author={Tanveer, Maham and Zhou, Yang and Niklaus, Simon and Amiri, Ali Mahdavi and Zhang, Hao and Singh, Krishna Kumar and Zhao, Nanxuan},
  journal={arXiv preprint arXiv:2412.13190},
  year={2024}
}

@inproceedings{denton2017unsupervised,
  title={Unsupervised learning of disentangled representations from video},
  author={Denton, Emily L and others},
  booktitle=NIPS,
  year={2017}
}

@inproceedings{liu2021infinite,
  title={Infinite nature: Perpetual view generation of natural scenes from a single image},
  author={Liu, Andrew and Tucker, Richard and Jampani, Varun and Makadia, Ameesh and Snavely, Noah and Kanazawa, Angjoo},
  booktitle=ICCV,
  year={2021}
}

@inproceedings{tulyakov2018mocogan,
  title={Mocogan: Decomposing motion and content for video generation},
  author={Tulyakov, Sergey and Liu, Ming-Yu and Yang, Xiaodong and Kautz, Jan},
  booktitle=CVPR,
  year={2018}
}

@inproceedings{li2022infinitenature,
  title={Infinitenature-zero: Learning perpetual view generation of natural scenes from single images},
  author={Li, Zhengqi and Wang, Qianqian and Snavely, Noah and Kanazawa, Angjoo},
  booktitle=ECCV,
  year={2022}
}

@inproceedings{villegas2017decomposing,
  title={Decomposing motion and content for natural video sequence prediction},
  author={Villegas, Ruben and Yang, Jimei and Hong, Seunghoon and Lin, Xunyu and Lee, Honglak},
  booktitle=ICLR,
  year={2017}
}

@article{vondrick2016generating,
  title={Generating videos with scene dynamics},
  author={Vondrick, Carl and Pirsiavash, Hamed and Torralba, Antonio},
  journal=NIPS,
  year={2016}
}

@inproceedings{gupta2024photorealistic,
  title={Photorealistic video generation with diffusion models},
  author={Gupta, Agrim and Yu, Lijun and Sohn, Kihyuk and Gu, Xiuye and Hahn, Meera and Li, Fei-Fei and Essa, Irfan and Jiang, Lu and Lezama, Jos{\'e}},
  booktitle=ECCV,
  year={2024}
}

@inproceedings{weissenborn2020scaling,
  title={Scaling Autoregressive Video Models},
  author={Weissenborn, Dirk and T{\"a}ckstr{\"o}m, Oscar and Uszkoreit, Jakob},
  booktitle=ICLR,
  year={2020}
}

@inproceedings{kondratyuk2024videopoet,
  title={VideoPoet: A Large Language Model for Zero-Shot Video Generation},
  author={Kondratyuk, Dan and Yu, Lijun and Gu, Xiuye and Lezama, Jose and Huang, Jonathan and Schindler, Grant and Hornung, Rachel and Birodkar, Vighnesh and Yan, Jimmy and Chiu, Ming-Chang and others},
  booktitle=ICML,
  year={2024}
}

@article{yan2021videogpt,
  title={Videogpt: Video generation using vq-vae and transformers},
  author={Yan, Wilson and Zhang, Yunzhi and Abbeel, Pieter and Srinivas, Aravind},
  journal={arXiv preprint arXiv:2104.10157},
  year={2021}
}

@article{ren2025next,
  title={Next Block Prediction: Video Generation via Semi-Auto-Regressive Modeling},
  author={Ren, Shuhuai and Ma, Shuming and Sun, Xu and Wei, Furu},
  journal={arXiv preprint arXiv:2502.07737},
  year={2025}
}

@inproceedings{bar2025navigation,
  title={Navigation world models},
  author={Bar, Amir and Zhou, Gaoyue and Tran, Danny and Darrell, Trevor and LeCun, Yann},
  booktitle={Proceedings of the Computer Vision and Pattern Recognition Conference},
  pages={15791--15801},
  year={2025}
}

@article{parkerholder2024genie2,
  title         = {Genie 2: A Large-Scale Foundation World Model},
  author        = {Jack Parker-Holder and Philip Ball and Jake Bruce and Vibhavari Dasagi and Kristian Holsheimer and Christos Kaplanis and Alexandre Moufarek and Guy Scully and Jeremy Shar and Jimmy Shi and Stephen Spencer and Jessica Yung and Michael Dennis and Sultan Kenjeyev and Shangbang Long and Vlad Mnih and Harris Chan and Maxime Gazeau and Bonnie Li and Fabio Pardo and Luyu Wang and Lei Zhang and Frederic Besse and Tim Harley and Anna Mitenkova and Jane Wang and Jeff Clune and Demis Hassabis and Raia Hadsell and Adrian Bolton and Satinder Singh and Tim Rockt{\"a}schel},
  year          = {2024},
  url           = {https://deepmind.google/discover/blog/genie-2-a-large-scale-foundation-world-model/}
}

@article{yang2024playable,
  title={Playable game generation},
  author={Yang, Mingyu and Li, Junyou and Fang, Zhongbin and Chen, Sheng and Yu, Yangbin and Fu, Qiang and Yang, Wei and Ye, Deheng},
  journal={arXiv preprint arXiv:2412.00887},
  year={2024}
}

@inproceedings{Bruce2024GenieGI,
  title={Genie: Generative interactive environments},
  author={Bruce, Jake and Dennis, Michael D and Edwards, Ashley and Parker-Holder, Jack and Shi, Yuge and Hughes, Edward and Lai, Matthew and Mavalankar, Aditi and Steigerwald, Richie and Apps, Chris and others},
  booktitle=ICML,
  year={2024}
}

@article{guo2025mineworld,
  title={Mineworld: a real-time and open-source interactive world model on minecraft},
  author={Guo, Junliang and Ye, Yang and He, Tianyu and Wu, Haoyu and Jiang, Yushu and Pearce, Tim and Bian, Jiang},
  journal={arXiv preprint arXiv:2504.08388},
  year={2025}
}

@article{po2025long,
  title={Long-context state-space video world models},
  author={Po, Ryan and Nitzan, Yotam and Zhang, Richard and Chen, Berlin and Dao, Tri and Shechtman, Eli and Wetzstein, Gordon and Huang, Xun},
  journal=ICCV,
  year={2025}
}

@article{yu2025gamefactory,
  title={Gamefactory: Creating new games with generative interactive videos},
  author={Yu, Jiwen and Qin, Yiran and Wang, Xintao and Wan, Pengfei and Zhang, Di and Liu, Xihui},
  journal={arXiv preprint arXiv:2501.08325},
  year={2025}
}

@article{he2025matrix,
  title={Matrix-Game 2.0: An Open-Source, Real-Time, and Streaming Interactive World Model},
  author={He, Xianglong and Peng, Chunli and Liu, Zexiang and Wang, Boyang and Zhang, Yifan and Cui, Qi and Kang, Fei and Jiang, Biao and An, Mengyin and Ren, Yangyang and others},
  journal={arXiv preprint arXiv:2508.13009},
  year={2025}
}

@article{xiao2023efficient,
  title={Efficient streaming language models with attention sinks},
  author={Xiao, Guangxuan and Tian, Yuandong and Chen, Beidi and Han, Song and Lewis, Mike},
  journal={arXiv preprint arXiv:2309.17453},
  year={2023}
}

@article{wan2025,
      title={Wan: Open and Advanced Large-Scale Video Generative Models}, 
      author={{Team Wan} and Ang Wang and Baole Ai and Bin Wen and Chaojie Mao and Chen-Wei Xie and Di Chen and Feiwu Yu and Haiming Zhao and Jianxiao Yang and Jianyuan Zeng and Jiayu Wang and Jingfeng Zhang and Jingren Zhou and Jinkai Wang and Jixuan Chen and Kai Zhu and Kang Zhao and Keyu Yan and Lianghua Huang and Mengyang Feng and Ningyi Zhang and Pandeng Li and Pingyu Wu and Ruihang Chu and Ruili Feng and Shiwei Zhang and Siyang Sun and Tao Fang and Tianxing Wang and Tianyi Gui and Tingyu Weng and Tong Shen and Wei Lin and Wei Wang and Wei Wang and Wenmeng Zhou and Wente Wang and Wenting Shen and Wenyuan Yu and Xianzhong Shi and Xiaoming Huang and Xin Xu and Yan Kou and Yangyu Lv and Yifei Li and Yijing Liu and Yiming Wang and Yingya Zhang and Yitong Huang and Yong Li and You Wu and Yu Liu and Yulin Pan and Yun Zheng and Yuntao Hong and Yupeng Shi and Yutong Feng and Zeyinzi Jiang and Zhen Han and Zhi-Fan Wu and Ziyu Liu},
      journal = {arXiv preprint arXiv:2503.20314},
      year={2025}
}

@article{low2025talkingmachines,
  title={TalkingMachines: Real-Time Audio-Driven FaceTime-Style Video via Autoregressive Diffusion Models},
  author={Low, Chetwin and Wang, Weimin},
  journal={arXiv preprint arXiv:2506.03099},
  year={2025}
}

@inproceedings{yin2024one,
  title={One-step diffusion with distribution matching distillation},
  author={Yin, Tianwei and Gharbi, Micha{\"e}l and Zhang, Richard and Shechtman, Eli and Durand, Fredo and Freeman, William T and Park, Taesung},
  booktitle={Proceedings of the IEEE/CVF conference on computer vision and pattern recognition},
  pages={6613--6623},
  year={2024}
}

@article{wang2024loong,
  title={Loong: Generating minute-level long videos with autoregressive language models},
  author={Wang, Yuqing and Xiong, Tianwei and Zhou, Daquan and Lin, Zhijie and Zhao, Yang and Kang, Bingyi and Feng, Jiashi and Liu, Xihui},
  journal={arXiv preprint arXiv:2410.02757},
  year={2024}
}

@article{Liu2022FlowSA,
  title={Flow Straight and Fast: Learning to Generate and Transfer Data with Rectified Flow},
  author={Xingchao Liu and Chengyue Gong and Qiang Liu},
  journal={ArXiv},
  year={2022},
  volume={abs/2209.03003},
  url={https://api.semanticscholar.org/CorpusID:252111177}
}

@article{nan2024openvid,
  title={Openvid-1m: A large-scale high-quality dataset for text-to-video generation},
  author={Nan, Kepan and Xie, Rui and Zhou, Penghao and Fan, Tiehan and Yang, Zhenheng and Chen, Zhijie and Li, Xiang and Yang, Jian and Tai, Ying},
  journal={arXiv preprint arXiv:2407.02371},
  year={2024}
}

@article{karaev2024cotracker3,
  title={Cotracker3: Simpler and better point tracking by pseudo-labelling real videos},
  author={Karaev, Nikita and Makarov, Iurii and Wang, Jianyuan and Neverova, Natalia and Vedaldi, Andrea and Rupprecht, Christian},
  journal={arXiv preprint arXiv:2410.11831},
  year={2024}
}

@article{yin2024improved,
  title={Improved distribution matching distillation for fast image synthesis},
  author={Yin, Tianwei and Gharbi, Micha{\"e}l and Park, Taesung and Zhang, Richard and Shechtman, Eli and Durand, Fredo and Freeman, Bill},
  journal={Advances in neural information processing systems},
  volume={37},
  pages={47455--47487},
  year={2024}
}

@article{zhang2025packing,
  title={Packing Input Frame Context in Next-Frame Prediction Models for Video Generation},
  author={Zhang, Lvmin and Agrawala, Maneesh},
  journal={arXiv preprint arXiv:2504.12626},
  year={2025}
}

@inproceedings{ruhe2024rolling,
  title={Rolling diffusion models},
  author={Ruhe, David and Heek, Jonathan and Salimans, Tim and Hoogeboom, Emiel},
  booktitle=ICML,
  year={2024}
}

@article{gu2025long,
  title={Long-Context Autoregressive Video Modeling with Next-Frame Prediction},
  author={Gu, Yuchao and Mao, Weijia and Shou, Mike Zheng},
  journal={arXiv preprint arXiv:2503.19325},
  year={2025}
}

@article{kim2024fifo,
  title={Fifo-diffusion: Generating infinite videos from text without training},
  author={Kim, Jihwan and Kang, Junoh and Choi, Jinyoung and Han, Bohyung},
  journal={Advances in Neural Information Processing Systems},
  volume={37},
  pages={89834--89868},
  year={2024}
}

@article{xie2024progressive,
  title={Progressive autoregressive video diffusion models},
  author={Xie, Desai and Xu, Zhan and Hong, Yicong and Tan, Hao and Liu, Difan and Liu, Feng and Kaufman, Arie and Zhou, Yang},
  journal={arXiv preprint arXiv:2410.08151},
  year={2024}
}

@inproceedings{deng2024nova,
  title={Autoregressive Video Generation without Vector Quantization},
  author={Deng, Haoge and Pan, Ting and Diao, Haiwen and Luo, Zhengxiong and Cui, Yufeng and Lu, Huchuan and Shan, Shiguang and Qi, Yonggang and Wang, Xinlong},
  booktitle=ICLR,
  year={2025}
}

@inproceedings{villegas2023phenaki,
  title={Phenaki: Variable Length Video Generation from Open Domain Textual Descriptions},
  author={Villegas, R and Moraldo, H and Castro, S and Babaeizadeh, M and Zhang, H and Kunze, J and Kindermans, PJ and Saffar, MT and Erhan, D},
  booktitle=ICLR,
  year={2023}
}

@article{blattmann2023stable,
  title={Stable video diffusion: Scaling latent video diffusion models to large datasets},
  author={Blattmann, Andreas and Dockhorn, Tim and Kulal, Sumith and Mendelevitch, Daniel and Kilian, Maciej and Lorenz, Dominik and Levi, Yam and English, Zion and Voleti, Vikram and Letts, Adam and others},
  journal={arXiv preprint arXiv:2311.15127},
  year={2023}
}

@inproceedings{yang2025cogvideox,
  title={Cogvideox: Text-to-video diffusion models with an expert transformer},
  author={Yang, Zhuoyi and Teng, Jiayan and Zheng, Wendi and Ding, Ming and Huang, Shiyu and Xu, Jiazheng and Yang, Yuanming and Hong, Wenyi and Zhang, Xiaohan and Feng, Guanyu and others},
  booktitle=ICLR,
  year={2025}
}

@article{polyak2024movie,
  title={Movie gen: A cast of media foundation models},
  author={Polyak, Adam and Zohar, Amit and Brown, Andrew and Tjandra, Andros and Sinha, Animesh and Lee, Ann and Vyas, Apoorv and Shi, Bowen and Ma, Chih-Yao and Chuang, Ching-Yao and others},
  journal={arXiv preprint arXiv:2410.13720},
  year={2024}
}

@article{kong2024hunyuanvideo,
  title={Hunyuanvideo: A systematic framework for large video generative models},
  author={Kong, Weijie and Tian, Qi and Zhang, Zijian and Min, Rox and Dai, Zuozhuo and Zhou, Jin and Xiong, Jiangfeng and Li, Xin and Wu, Bo and Zhang, Jianwei and others},
  journal={arXiv preprint arXiv:2412.03603},
  year={2024}
}

@misc{videoworldsimulators2024,
  title={Video generation models as world simulators},
  author={Tim Brooks and Bill Peebles and Connor Holmes and Will DePue and Yufei Guo and Li Jing and David Schnurr and Joe Taylor and Troy Luhman and Eric Luhman and Clarence Ng and Ricky Wang and Aditya Ramesh},
  year={2024},
  url={https://openai.com/research/video-generation-models-as-world-simulators},
}

@article{liu2024redefining,
  title={Redefining Temporal Modeling in Video Diffusion: The Vectorized Timestep Approach},
  author={Liu, Yaofang and Ren, Yumeng and Cun, Xiaodong and Artola, Aitor and Liu, Yang and Zeng, Tieyong and Chan, Raymond H and Morel, Jean-michel},
  journal={arXiv preprint arXiv:2410.03160},
  year={2024}
}

@article{team2025yan,
  title={Yan: Foundational Interactive Video Generation},
  author={{Yan Team}},
  journal={arXiv preprint arXiv:2508.08601},
  year={2025}
}

@article{li2025hunyuan,
  title={Hunyuan-GameCraft: High-dynamic Interactive Game Video Generation with Hybrid History Condition},
  author={Li, Jiaqi and Tang, Junshu and Xu, Zhiyong and Wu, Longhuang and Zhou, Yuan and Shao, Shuai and Yu, Tianbao and Cao, Zhiguo and Lu, Qinglin},
  journal={arXiv preprint arXiv:2506.17201},
  year={2025}
}

@misc{ball2025genie,
  title={Genie 3: A new frontier for world models},
  author={Ball, Philip J and Bauer, J and Belletti, F and others},
  year={2025}
}

@inproceedings{chen2024diffusion,
  title={Diffusion forcing: Next-token prediction meets full-sequence diffusion},
  author={Chen, Boyuan and Mart{\'\i} Mons{\'o}, Diego and Du, Yilun and Simchowitz, Max and Tedrake, Russ and Sitzmann, Vincent},
  booktitle=NIPS,
  year={2024}
}

@article{guo2025long,
  title={Long context tuning for video generation},
  author={Guo, Yuwei and Yang, Ceyuan and Yang, Ziyan and Ma, Zhibei and Lin, Zhijie and Yang, Zhenheng and Lin, Dahua and Jiang, Lu},
  journal={arXiv preprint arXiv:2503.10589},
  year={2025}
}

@inproceedings{zhang2018unreasonable,
  title={The unreasonable effectiveness of deep features as a perceptual metric},
  author={Zhang, Richard and Isola, Phillip and Efros, Alexei A and Shechtman, Eli and Wang, Oliver},
  booktitle={Proceedings of the IEEE conference on computer vision and pattern recognition},
  pages={586--595},
  year={2018}
}

@inproceedings{Perazzi2016,
  author = {F. Perazzi and J. Pont-Tuset and B. McWilliams and L. {Van Gool} and M. Gross and A. Sorkine-Hornung},
  title = {A Benchmark Dataset and Evaluation Methodology for Video Object Segmentation},
  booktitle = {Computer Vision and Pattern Recognition},
  year = {2016}
}

@inproceedings{weng2024art,
  title={Art-v: Auto-regressive text-to-video generation with diffusion models},
  author={Weng, Wenming and Feng, Ruoyu and Wang, Yanhui and Dai, Qi and Wang, Chunyu and Yin, Dacheng and Zhao, Zhiyuan and Qiu, Kai and Bao, Jianmin and Yuan, Yuhui and others},
  booktitle=CVPR,
  year={2024}
}

@inproceedings{yin2025causvid,
  title={From Slow Bidirectional to Fast Autoregressive Video Diffusion Models},
  author={Yin, Tianwei and Zhang, Qiang and Zhang, Richard and Freeman, William T and Durand, Fredo and Shechtman, Eli and Huang, Xun},
  booktitle=CVPR,
  year={2025}
}

@inproceedings{sun2025ar,
  title={AR-Diffusion: Asynchronous Video Generation with Auto-Regressive Diffusion},
  author={Sun, Mingzhen and Wang, Weining and Li, Gen and Liu, Jiawei and Sun, Jiahui and Feng, Wanquan and Lao, Shanshan and Zhou, SiYu and He, Qian and Liu, Jing},
  booktitle=CVPR,
  year={2025}
}

@article{hu2024acdit,
  title={ACDiT: Interpolating Autoregressive Conditional Modeling and Diffusion Transformer},
  author={Hu, Jinyi and Hu, Shengding and Song, Yuxuan and Huang, Yufei and Wang, Mingxuan and Zhou, Hao and Liu, Zhiyuan and Ma, Wei-Ying and Sun, Maosong},
  journal={arXiv preprint arXiv:2412.07720},
  year={2024}
}

@inproceedings{jin2024pyramidal,
  title={Pyramidal Flow Matching for Efficient Video Generative Modeling},
  author={Jin, Yang and Sun, Zhicheng and Li, Ningyuan and Xu, Kun and Jiang, Hao and Zhuang, Nan and Huang, Quzhe and Song, Yang and Mu, Yadong and Lin, Zhouchen},
  booktitle=ICLR,
  year={2025}
}

@inproceedings{ho2022video,
  title={Video diffusion models},
  author={Ho, Jonathan and Salimans, Tim and Gritsenko, Alexey and Chan, William and Norouzi, Mohammad and Fleet, David J},
  booktitle=NIPS,
  year={2022}
}

@article{gao2024ca2,
  title={Ca2-VDM: Efficient Autoregressive Video Diffusion Model with Causal Generation and Cache Sharing},
  author={Gao, Kaifeng and Shi, Jiaxin and Zhang, Hanwang and Wang, Chunping and Xiao, Jun and Chen, Long},
  journal={arXiv preprint arXiv:2411.16375},
  year={2024}
}

@inproceedings{li2025arlon,
  title={Arlon: Boosting diffusion transformers with autoregressive models for long video generation},
  author={Li, Zongyi and Hu, Shujie and Liu, Shujie and Zhou, Long and Choi, Jeongsoo and Meng, Lingwei and Guo, Xun and Li, Jinyu and Ling, Hefei and Wei, Furu},
  booktitle=ICLR,
  year={2025}
}

@inproceedings{blattmann2023align,
  title={Align your latents: High-resolution video synthesis with latent diffusion models},
  author={Blattmann, Andreas and Rombach, Robin and Ling, Huan and Dockhorn, Tim and Kim, Seung Wook and Fidler, Sanja and Kreis, Karsten},
  booktitle=CVPR,
  year={2023}
}

@article{brooks2022generating,
  title={Generating long videos of dynamic scenes},
  author={Brooks, Tim and Hellsten, Janne and Aittala, Miika and Wang, Ting-Chun and Aila, Timo and Lehtinen, Jaakko and Liu, Ming-Yu and Efros, Alexei and Karras, Tero},
  journal=NIPS,
  year={2022}
}

@article{zhang2025test,
  title={Test-Time Training Done Right},
  author={Zhang, Tianyuan and Bi, Sai and Hong, Yicong and Zhang, Kai and Luan, Fujun and Yang, Songlin and Sunkavalli, Kalyan and Freeman, William T and Tan, Hao},
  journal={arXiv preprint arXiv:2505.23884},
  year={2025}
}

@article{lin2025diffusion,
  title={Diffusion adversarial post-training for one-step video generation},
  author={Lin, Shanchuan and Xia, Xin and Ren, Yuxi and Yang, Ceyuan and Xiao, Xuefeng and Jiang, Lu},
  journal={arXiv preprint arXiv:2501.08316},
  year={2025}
}

@article{huang2025self,
  title={Self Forcing: Bridging the Train-Test Gap in Autoregressive Video Diffusion},
  author={Huang, Xun and Li, Zhengqi and He, Guande and Zhou, Mingyuan and Shechtman, Eli},
  journal={arXiv preprint arXiv:2506.08009},
  year={2025}
}

@article{pang2024dreamdance,
  title={Dreamdance: Animating human images by enriching 3d geometry cues from 2d poses},
  author={Pang, Yatian and Zhu, Bin and Lin, Bin and Zheng, Mingzhe and Tay, Francis EH and Lim, Ser-Nam and Yang, Harry and Yuan, Li},
  journal={arXiv preprint arXiv:2412.00397},
  year={2024}
}

@article{jiang2025vidsketch,
  title={VidSketch: Hand-drawn Sketch-Driven Video Generation with Diffusion Control},
  author={Jiang, Lifan and Chen, Shuang and Wu, Boxi and Guan, Xiaotong and Zhang, Jiahui},
  journal={arXiv preprint arXiv:2502.01101},
  year={2025}
}

@article{yang2025layeranimate,
  title={LayerAnimate: Layer-level Control for Animation},
  author={Yang, Yuxue and Fan, Lue and Lin, Zuzeng and Wang, Feng and Zhang, Zhaoxiang},
  journal={arXiv preprint arXiv:2501.08295},
  year={2025}
}

@article{wu2024motionbooth,
  title={Motionbooth: Motion-aware customized text-to-video generation},
  author={Wu, Jianzong and Li, Xiangtai and Zeng, Yanhong and Zhang, Jiangning and Zhou, Qianyu and Li, Yining and Tong, Yunhai and Chen, Kai},
  journal={Advances in Neural Information Processing Systems},
  volume={37},
  pages={34322--34348},
  year={2024}
}

@article{gillman2025force,
  title={Force Prompting: Video Generation Models Can Learn and Generalize Physics-based Control Signals},
  author={Gillman, Nate and Herrmann, Charles and Freeman, Michael and Aggarwal, Daksh and Luo, Evan and Sun, Deqing and Sun, Chen},
  journal={arXiv preprint arXiv:2505.19386},
  year={2025}
}

@article{lipman2022flow,
  title={Flow matching for generative modeling},
  author={Lipman, Yaron and Chen, Ricky TQ and Ben-Hamu, Heli and Nickel, Maximilian and Le, Matt},
  journal={arXiv preprint arXiv:2210.02747},
  year={2022}
}

@inproceedings{zhang2023adding,
  title={Adding conditional control to text-to-image diffusion models},
  author={Zhang, Lvmin and Rao, Anyi and Agrawala, Maneesh},
  booktitle={Proceedings of the IEEE/CVF international conference on computer vision},
  pages={3836--3847},
  year={2023}
}

@inproceedings{wang2024motionctrl,
  title={Motionctrl: A unified and flexible motion controller for video generation},
  author={Wang, Zhouxia and Yuan, Ziyang and Wang, Xintao and Li, Yaowei and Chen, Tianshui and Xia, Menghan and Luo, Ping and Shan, Ying},
  booktitle={SIGGRAPH},
  year={2024}
}

@inproceedings{geng2025motion,
  title={Motion prompting: Controlling video generation with motion trajectories},
  author={Geng, Daniel and Herrmann, Charles and Hur, Junhwa and Cole, Forrester and Zhang, Serena and Pfaff, Tobias and Lopez-Guevara, Tatiana and Aytar, Yusuf and Rubinstein, Michael and Sun, Chen and others},
  booktitle=CVPR,
  year={2025}
}

@inproceedings{burgert2025go,
  title={Go-with-the-flow: Motion-controllable video diffusion models using real-time warped noise},
  author={Burgert, Ryan and Xu, Yuancheng and Xian, Wenqi and Pilarski, Oliver and Clausen, Pascal and He, Mingming and Ma, Li and Deng, Yitong and Li, Lingxiao and Mousavi, Mohsen and others},
  booktitle=CVPR,
  year={2025}
}

@inproceedings{zhang2025tora,
  title={Tora: Trajectory-oriented diffusion transformer for video generation},
  author={Zhang, Zhenghao and Liao, Junchao and Li, Menghao and Dai, Zuozhuo and Qiu, Bingxue and Zhu, Siyu and Qin, Long and Wang, Weizhi},
  booktitle=CVPR,
  year={2025}
}

@inproceedings{li2025magicmotion,
  title={Magicmotion: Controllable video generation with dense-to-sparse trajectory guidance},
  author={Li, Quanhao and Xing, Zhen and Wang, Rui and Zhang, Hui and Dai, Qi and Wu, Zuxuan},
  booktitle=ICCV,
  year={2025}
}

@article{wang2025ati,
  title={ATI: Any Trajectory Instruction for Controllable Video Generation},
  author={Wang, Angtian and Huang, Haibin and Fang, Jacob Zhiyuan and Yang, Yiding and Ma, Chongyang},
  journal={arXiv preprint arXiv:2505.22944},
  year={2025}
}

@inproceedings{li2024generative,
  title={Generative image dynamics},
  author={Li, Zhengqi and Tucker, Richard and Snavely, Noah and Holynski, Aleksander},
  booktitle={Proceedings of the IEEE/CVF Conference on Computer Vision and Pattern Recognition},
  pages={24142--24153},
  year={2024}
}

@inproceedings{bahmani20244d,
  title={4d-fy: Text-to-4d generation using hybrid score distillation sampling},
  author={Bahmani, Sherwin and Skorokhodov, Ivan and Rong, Victor and Wetzstein, Gordon and Guibas, Leonidas and Wonka, Peter and Tulyakov, Sergey and Park, Jeong Joon and Tagliasacchi, Andrea and Lindell, David B},
  booktitle={Proceedings of the IEEE/CVF Conference on Computer Vision and Pattern Recognition},
  pages={7996--8006},
  year={2024}
}

@article{xing2024tooncrafter,
  title={Tooncrafter: Generative cartoon interpolation},
  author={Xing, Jinbo and Liu, Hanyuan and Xia, Menghan and Zhang, Yong and Wang, Xintao and Shan, Ying and Wong, Tien-Tsin},
  journal={ACM Transactions on Graphics (TOG)},
  volume={43},
  number={6},
  pages={1--11},
  year={2024},
  publisher={ACM New York, NY, USA}
}

@article{fu2025learning,
  title={Learning Video Generation for Robotic Manipulation with Collaborative Trajectory Control},
  author={Fu, Xiao and Wang, Xintao and Liu, Xian and Bai, Jianhong and Xu, Runsen and Wan, Pengfei and Zhang, Di and Lin, Dahua},
  journal={arXiv preprint arXiv:2506.01943},
  year={2025}
}

@article{wu2025video,
  title={Video World Models with Long-term Spatial Memory},
  author={Wu, Tong and Yang, Shuai and Po, Ryan and Xu, Yinghao and Liu, Ziwei and Lin, Dahua and Wetzstein, Gordon},
  journal={arXiv preprint arXiv:2506.05284},
  year={2025}
}

@article{gao2024vista,
  title={Vista: A generalizable driving world model with high fidelity and versatile controllability},
  author={Gao, Shenyuan and Yang, Jiazhi and Chen, Li and Chitta, Kashyap and Qiu, Yihang and Geiger, Andreas and Zhang, Jun and Li, Hongyang},
  journal={Advances in Neural Information Processing Systems},
  volume={37},
  pages={91560--91596},
  year={2024}
}

@inproceedings{tu2025videoanydoor,
  title={Videoanydoor: High-fidelity video object insertion with precise motion control},
  author={Tu, Yuanpeng and Luo, Hao and Chen, Xi and Ji, Sihui and Bai, Xiang and Zhao, Hengshuang},
  booktitle={Proceedings of the Special Interest Group on Computer Graphics and Interactive Techniques Conference Conference Papers},
  pages={1--11},
  year={2025}
}

@article{li2025diffueraser,
  title={Diffueraser: A diffusion model for video inpainting},
  author={Li, Xiaowen and Xue, Haolan and Ren, Peiran and Bo, Liefeng},
  journal={arXiv preprint arXiv:2501.10018},
  year={2025}
}

@article{zheng2025vidcraft3,
  title={Vidcraft3: Camera, object, and lighting control for image-to-video generation},
  author={Zheng, Sixiao and Peng, Zimian and Zhou, Yanpeng and Zhu, Yi and Xu, Hang and Huang, Xiangru and Fu, Yanwei},
  journal={arXiv preprint arXiv:2502.07531},
  year={2025}
}

@inproceedings{li2025puppet,
  title={Puppet-master: Scaling interactive video generation as a motion prior for part-level dynamics},
  author={Li, Ruining and Zheng, Chuanxia and Rupprecht, Christian and Vedaldi, Andrea},
  booktitle={Proceedings of the IEEE/CVF International Conference on Computer Vision},
  pages={13405--13415},
  year={2025}
}

@inproceedings{yang2024direct,
  title={Direct-a-video: Customized video generation with user-directed camera movement and object motion},
  author={Yang, Shiyuan and Hou, Liang and Huang, Haibin and Ma, Chongyang and Wan, Pengfei and Zhang, Di and Chen, Xiaodong and Liao, Jing},
  booktitle={ACM SIGGRAPH 2024 Conference Papers},
  pages={1--12},
  year={2024}
}

@article{bahmani2024vd3d,
  title={Vd3d: Taming large video diffusion transformers for 3d camera control},
  author={Bahmani, Sherwin and Skorokhodov, Ivan and Siarohin, Aliaksandr and Menapace, Willi and Qian, Guocheng and Vasilkovsky, Michael and Lee, Hsin-Ying and Wang, Chaoyang and Zou, Jiaxu and Tagliasacchi, Andrea and others},
  journal={arXiv preprint arXiv:2407.12781},
  year={2024}
}

@article{gao2024cat3d,
  title={Cat3d: Create anything in 3d with multi-view diffusion models},
  author={Gao, Ruiqi and Holynski, Aleksander and Henzler, Philipp and Brussee, Arthur and Martin-Brualla, Ricardo and Srinivasan, Pratul and Barron, Jonathan T and Poole, Ben},
  journal={arXiv preprint arXiv:2405.10314},
  year={2024}
}

@inproceedings{wu2025cat4d,
  title={Cat4d: Create anything in 4d with multi-view video diffusion models},
  author={Wu, Rundi and Gao, Ruiqi and Poole, Ben and Trevithick, Alex and Zheng, Changxi and Barron, Jonathan T and Holynski, Aleksander},
  booktitle={Proceedings of the Computer Vision and Pattern Recognition Conference},
  pages={26057--26068},
  year={2025}
}

@article{bai2025recammaster,
  title={Recammaster: Camera-controlled generative rendering from a single video},
  author={Bai, Jianhong and Xia, Menghan and Fu, Xiao and Wang, Xintao and Mu, Lianrui and Cao, Jinwen and Liu, Zuozhu and Hu, Haoji and Bai, Xiang and Wan, Pengfei and others},
  journal={arXiv preprint arXiv:2503.11647},
  year={2025}
}

@article{zheng2024cami2v,
  title={Cami2v: Camera-controlled image-to-video diffusion model},
  author={Zheng, Guangcong and Li, Teng and Jiang, Rui and Lu, Yehao and Wu, Tao and Li, Xi},
  journal={arXiv preprint arXiv:2410.15957},
  year={2024}
}

@article{he2024cameractrl,
  title={Cameractrl: Enabling camera control for text-to-video generation},
  author={He, Hao and Xu, Yinghao and Guo, Yuwei and Wetzstein, Gordon and Dai, Bo and Li, Hongsheng and Yang, Ceyuan},
  journal={arXiv preprint arXiv:2404.02101},
  year={2024}
}

@inproceedings{schonberger2016structure,
  title={Structure-from-motion revisited},
  author={Schonberger, Johannes L and Frahm, Jan-Michael},
  booktitle={Proceedings of the IEEE conference on computer vision and pattern recognition},
  pages={4104--4113},
  year={2016}
}

@article{wang2025moge,
  title={MoGe-2: Accurate Monocular Geometry with Metric Scale and Sharp Details},
  author={Wang, Ruicheng and Xu, Sicheng and Dong, Yue and Deng, Yu and Xiang, Jianfeng and Lv, Zelong and Sun, Guangzhong and Tong, Xin and Yang, Jiaolong},
  journal={arXiv preprint arXiv:2507.02546},
  year={2025}
}

@article{mildenhall2019local,
  title={Local light field fusion: Practical view synthesis with prescriptive sampling guidelines},
  author={Mildenhall, Ben and Srinivasan, Pratul P and Ortiz-Cayon, Rodrigo and Kalantari, Nima Khademi and Ramamoorthi, Ravi and Ng, Ren and Kar, Abhishek},
  journal={ACM Transactions on Graphics (ToG)},
  volume={38},
  number={4},
  pages={1--14},
  year={2019},
  publisher={ACM New York, NY, USA}
}

@inproceedings{niu2024mofa,
  title={Mofa-video: Controllable image animation via generative motion field adaptions in frozen image-to-video diffusion model},
  author={Niu, Muyao and Cun, Xiaodong and Wang, Xintao and Zhang, Yong and Shan, Ying and Zheng, Yinqiang},
  booktitle=ECCV,
  year={2024}
}

@inproceedings{xu2025depthsplat,
  title={Depthsplat: Connecting gaussian splatting and depth},
  author={Xu, Haofei and Peng, Songyou and Wang, Fangjinhua and Blum, Hermann and Barath, Daniel and Geiger, Andreas and Pollefeys, Marc},
  booktitle={Proceedings of the Computer Vision and Pattern Recognition Conference},
  pages={16453--16463},
  year={2025}
}

@article{yu2024viewcrafter,
  title={Viewcrafter: Taming video diffusion models for high-fidelity novel view synthesis},
  author={Yu, Wangbo and Xing, Jinbo and Yuan, Li and Hu, Wenbo and Li, Xiaoyu and Huang, Zhipeng and Gao, Xiangjun and Wong, Tien-Tsin and Shan, Ying and Tian, Yonghong},
  journal={arXiv preprint arXiv:2409.02048},
  year={2024}
}

@article{zhou2025stable,
  title={Stable virtual camera: Generative view synthesis with diffusion models},
  author={Zhou, Jensen Jinghao and Gao, Hang and Voleti, Vikram and Vasishta, Aaryaman and Yao, Chun-Han and Boss, Mark and Torr, Philip and Rupprecht, Christian and Jampani, Varun},
  journal={arXiv preprint arXiv:2503.14489},
  year={2025}
}

@inproceedings{shi2024motion,
  title={Motion-i2v: Consistent and controllable image-to-video generation with explicit motion modeling},
  author={Shi, Xiaoyu and Huang, Zhaoyang and Wang, Fu-Yun and Bian, Weikang and Li, Dasong and Zhang, Yi and Zhang, Manyuan and Cheung, Ka Chun and See, Simon and Qin, Hongwei and others},
  booktitle={ACM SIGGRAPH 2024 Conference Papers},
  pages={1--11},
  year={2024}
}

@inproceedings{zhou2025trackgo,
  title={Trackgo: A flexible and efficient method for controllable video generation},
  author={Zhou, Haitao and Wang, Chuang and Nie, Rui and Liu, Jinlin and Yu, Dongdong and Yu, Qian and Wang, Changhu},
  booktitle=AAAI,
  year={2025}
}

@inproceedings{lei2025animateanything,
  title={Animateanything: Consistent and controllable animation for video generation},
  author={Lei, Guojun and Wang, Chi and Zhang, Rong and Wang, Yikai and Li, Hong and Xu, Weiwei},
  booktitle=CVPR,
  year={2025}
}

@inproceedings{li2025image,
  title={Image conductor: Precision control for interactive video synthesis},
  author={Li, Yaowei and Wang, Xintao and Zhang, Zhaoyang and Wang, Zhouxia and Yuan, Ziyang and Xie, Liangbin and Shan, Ying and Zou, Yuexian},
  booktitle=AAAI,
  year={2025}
}

@article{wang2024vidprom,
  title={Vidprom: A million-scale real prompt-gallery dataset for text-to-video diffusion models},
  author={Wang, Wenhao and Yang, Yi},
  journal={Advances in Neural Information Processing Systems},
  volume={37},
  pages={65618--65642},
  year={2024}
}

@inproceedings{pan2023drag,
  title={Drag your gan: Interactive point-based manipulation on the generative image manifold},
  author={Pan, Xingang and Tewari, Ayush and Leimk{\"u}hler, Thomas and Liu, Lingjie and Meka, Abhimitra and Theobalt, Christian},
  booktitle={ACM SIGGRAPH 2023 conference proceedings},
  pages={1--11},
  year={2023}
}

@inproceedings{shi2024dragdiffusion,
  title={Dragdiffusion: Harnessing diffusion models for interactive point-based image editing},
  author={Shi, Yujun and Xue, Chuhui and Liew, Jun Hao and Pan, Jiachun and Yan, Hanshu and Zhang, Wenqing and Tan, Vincent YF and Bai, Song},
  booktitle={Proceedings of the IEEE/CVF Conference on Computer Vision and Pattern Recognition},
  pages={8839--8849},
  year={2024}
}

@article{choi2025enhancing,
  title={Enhancing Motion Dynamics of Image-to-Video Models via Adaptive Low-Pass Guidance},
  author={Choi, June Suk and Lee, Kyungmin and Yu, Sihyun and Choi, Yisol and Shin, Jinwoo and Lee, Kimin},
  journal={arXiv preprint arXiv:2506.08456},
  year={2025}
}

@article{nie2023blessing,
  title={The blessing of randomness: Sde beats ode in general diffusion-based image editing},
  author={Nie, Shen and Guo, Hanzhong Allan and Lu, Cheng and Zhou, Yuhao and Zheng, Chenyu and Li, Chongxuan},
  journal={arXiv preprint arXiv:2311.01410},
  year={2023}
}

@inproceedings{shin2024instantdrag,
  title={Instantdrag: Improving interactivity in drag-based image editing},
  author={Shin, Joonghyuk and Choi, Daehyeon and Park, Jaesik},
  booktitle={SIGGRAPH Asia 2024 Conference Papers},
  pages={1--10},
  year={2024}
}

@article{zhao2024fastdrag,
  title={Fastdrag: Manipulate anything in one step},
  author={Zhao, Xuanjia and Guan, Jian and Fan, Congyi and Xu, Dongli and Lin, Youtian and Pan, Haiwei and Feng, Pengming},
  journal={Advances in Neural Information Processing Systems},
  volume={37},
  pages={74439--74460},
  year={2024}
}
\bibliographystyle{iclr2026_conference}

\clearpage
\appendix
\setcounter{table}{0}
\setcounter{figure}{0}
\setcounter{equation}{0}
\renewcommand{\thefigure}{A\arabic{figure}}
\renewcommand{\thetable}{A\arabic{table}}
\renewcommand{\theequation}{A\arabic{equation}}

\section{Training Efficient Tiny VAE}
\label{sec:sup_tiny}
\begin{table}[h]
\centering
\scriptsize
\caption{\textbf{Comparison of Causal VAE Models.} We evaluate reconstruction quality on the Sora demo samples (resized to 81f$\times$832$\times$480) by encoding videos with the Full VAE encoder and decoding with different VAE variants. Our Tiny VAE achieves an order-of-magnitude faster decoding than Full VAEs while outperforming existing community implementations in reconstruction quality.}
\label{tab:vae_comparison}
\begin{tabular}{lcccccc}
\toprule
Model & \begin{tabular}[c]{@{}c@{}}Decoder\\Params\end{tabular} & \begin{tabular}[c]{@{}c@{}}Compression\\Rate\end{tabular} & \begin{tabular}[c]{@{}c@{}}Decoding Time(s)\\(81×832×480)\end{tabular} & PSNR & SSIM & LPIPS \\
\midrule
Full VAE (Wan 2.1) & 73.3M & 8×8×4 & 1.67 & 31.43 & 0.934 & 0.069 \\
Tiny VAE (Wan 2.1 ~\citep{BoerBohan2025TAEHV}) & 9.84M & 8×8×4 & 0.12 & 28.85 & 0.899 & 0.168 \\
Tiny VAE (Wan 2.1, Ours) & 9.84M & 8×8×4 & 0.12 & 29.27 & 0.904 & 0.107 \\
\midrule
Full VAE (Wan 2.2, 5B) & 555M & 16×16×4 & 1.75 & 31.87 & 0.938 & 0.065 \\
Tiny VAE (Wan 2.2, Ours) & 56.7M & 16×16×4 & 0.23 & 28.43 & 0.883 & 0.126 \\
\bottomrule
\end{tabular}
\end{table}
As discussed in the main paper, the full CausalVAE becomes a bottleneck in streaming pipelines. Wan 2.1's CausalVAE performs $8\times$ spatial and $4\times$ temporal compression, while Wan 2.2 performs $16\times$ spatial and $4\times$ temporal compression. VAE decoding using the full CausalVAE takes 47\% of chunk generation wall time for Wan 2.1 (1.3B), and 35\% for Wan 2.2 (5B) in our base setup. Although Wan 2.2 (5B) VAE's higher compression rate enables Wan 2.2 to generate 720P videos in real time, it increases the decoding time and memory footprint.

Inspired by community implementations of Tiny VAE~\citep{BoerBohan2025TAEHV}, we train a compact decoder from scratch with larger data, cleaner training pipelines, and better loss designs. While \citet{BoerBohan2025TAEHV} trains Tiny VAE by regressing outputs to the original Wan VAE using adversarial loss from a PatchGAN discriminator and reconstruction loss with replay buffer, we extend this design by incorporating LPIPS loss with proper data scaling, and hyperparameter selections. We use a random subset of videos from OpenVid-1M and synthetic Wan videos (total 280K samples), training for 200K steps with a learning rate of $3\times10^{-4}$, batch size of 16, and AdamW optimizer. We train at a lower resolution of $144\times144$ and frame length of 21, but the model scales well to larger video dimensions.

As shown in Table~\ref{tab:vae_comparison}, Tiny VAE achieves substantially faster decoding with significantly fewer parameters compared to the Full VAE. While Tiny VAEs typically produce slightly lower reconstruction quality than Full VAEs, our implementation substantially outperforms existing community versions. Importantly, when used jointly with our distilled student for latent decoding, we observe minimal quality differences in practice (Table~\ref{tab:vae_streaming}), as most quality degradation and drift originate from the diffusion model itself rather than VAE reconstruction. For consistency when evaluating the performance of distilled diffusion model, we use results from Full VAE and report corresponding speed, and adopt Tiny VAE for the streaming demo.

\begin{table}[h]
\centering
\caption{\textbf{Evaluating Tiny VAE in Streaming Generation Setup.} Using the same distilled student model, we ablate the impact of switching VAE from original Full VAE to Tiny VAE in Sora demo subset. It's important to note that even after changing to Tiny VAE, \textit{our distilled models still outperform all other baselines} and quality differences compared to Full VAEs are marginal while achieving 1.75$\times$ and 2.3$\times$ higher throughput.}
\label{tab:vae_streaming}
\small
\begin{tabular}{lcccccc}
\toprule
Model & Throughput (FPS) & Latency (s) & PSNR & SSIM & LPIPS \\
\midrule
Full VAE (Wan 2.1) & 16.7 & 0.69 & 16.67 & 0.531 & 0.360 \\
Tiny VAE (Wan 2.1, Ours) & 29.5 & 0.39 & 16.68 & 0.528 & 0.365 \\
\midrule
Full VAE (Wan 2.2, 5B) & 10.4 & 1.14 & 16.62 & 0.545 & 0.343 \\
Tiny VAE (Wan 2.2, Ours) & 23.9 & 0.49 & 16.62 & 0.543 & 0.349 \\
\bottomrule
\end{tabular}
\end{table}

\section{VBench results and User study}
We additionally evaluate \methodname using VBench-I2V~\citep{huang2024vbench} and conduct user studies on 20 samples from the Sora demo subset. In Vbench-I2V, we exclude camera motion and dynamic degree metrics since these dimensions are already constrained by the input trajectory conditions rather than text prompts. VBench results show strong correlation with the underlying backbone model, favoring recent Wan-based architectures. The provision of image and trajectory conditions leads to uniformly high scores across methods, reducing discriminative power between models. Nonetheless, both our teacher and distilled models consistently achieve competitive performance across all evaluated dimensions.

\begin{figure}[h]
\centering
\includegraphics[width=0.75\textwidth]{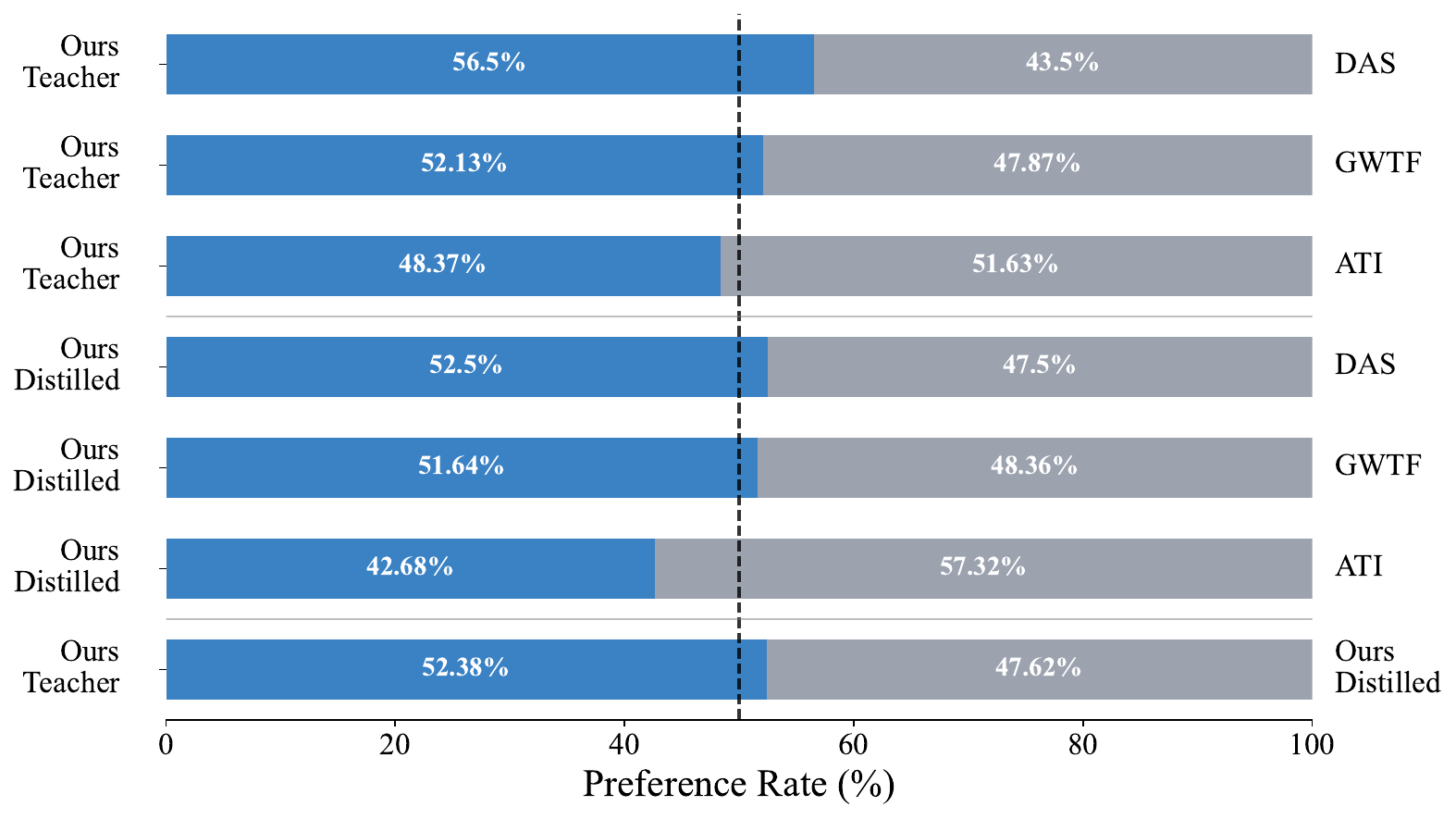}
\caption{\textbf{User Study Results.} We evaluate video quality through pairwise comparisons on 20 Sora samples. In terms of pure video quality, our models outperform all baselines except ATI, which uses a 10$\times$ larger backbone (Wan 2.1-14B), producing visually favorable videos.}
\vspace{-3mm}
\label{fig:user_study_results}
\end{figure}

For user study, we collected 2,800 responses evaluating video quality of generated videos using our Wan 2.1 (1.3B) variants. Since accurately assessing track-following capability from thousands of grid point trajectories is challenging for participants, we focused solely on video quality assessment, with results shown in Figure~\ref{fig:user_study_results}. As with VBench, video quality correlates strongly with backbone capacity. Notably, ATI~\citep{wang2025ati}, which leverages Wan 2.1 14B (10$\times$ larger than our 1.3B model), generally produces more visually favorable videos. However, despite ATI's aesthetic quality, we observe it often lacks precise trajectory adherence. Both our teacher and student models outperform other baselines in quality, with the teacher being slightly preferred over the student.

\begin{table}[t]
\scriptsize
\centering
\caption{\textbf{VBench-I2V Results.} We evaluate other baselines using VBench-I2V on Sora subset. While the results primarily depend on the choice of backbone, our models generally achieve high performance across all dimensions.}
\label{tab:i2v_comparison}
\begin{tabular}{lccccccc}
\toprule
Method & \begin{tabular}[c]{@{}c@{}}i2v\\subject\end{tabular} & \begin{tabular}[c]{@{}c@{}}i2v\\background\end{tabular} & \begin{tabular}[c]{@{}c@{}}subject\\consistency\end{tabular} & \begin{tabular}[c]{@{}c@{}}background\\consistency\end{tabular} & \begin{tabular}[c]{@{}c@{}}motion\\smoothness\end{tabular} & \begin{tabular}[c]{@{}c@{}}aesthetic\\quality\end{tabular} & \begin{tabular}[c]{@{}c@{}}imaging\\quality\end{tabular} \\
\midrule
Image Conductor~\citep{li2025image} & 0.847 & 0.868 & 0.791 & 0.889 & 0.906 & 0.505 & 0.689 \\
GWTF~\citep{burgert2025go} & 0.957 & 0.974 & 0.933 & 0.944 & 0.981 & 0.620 & 0.675 \\
DAS~\citep{gu2025diffusion} & 0.972 & 0.987 & \textbf{0.953} & \underline{0.958} & \textbf{0.988} & \underline{0.634} & 0.695 \\
ATI~\citep{wang2025ati} & 0.981 & \underline{0.988} & \underline{0.948} & 0.947 & 0.980 & 0.629 & \textbf{0.707} \\
\midrule
Ours Teacher (1.3B) & \textbf{0.984} & \underline{0.988} & \underline{0.948} & 0.943 & \underline{0.987} & 0.625 & 0.698 \\
Ours Distilled (1.3B) & 0.982 & 0.987 & 0.940 & 0.941 & 0.985 & 0.618 & 0.684 \\
Ours Teacher (5B) & \underline{0.983} & \underline{0.988} & 0.947 & \textbf{0.959} & 0.982 & \textbf{0.637} & \textbf{0.707} \\
Ours Distilled (5B) & \textbf{0.984} & \textbf{0.990} & 0.945 & \textbf{0.959} & \underline{0.987} & 0.630 & \underline{0.703} \\
\bottomrule
\end{tabular}
\end{table}

\section{Additional Ablation Experiments and Qualitative Results}
\label{sec:sup_abl}
\begin{figure}[t]
\centering
\includegraphics[width=0.75\textwidth]{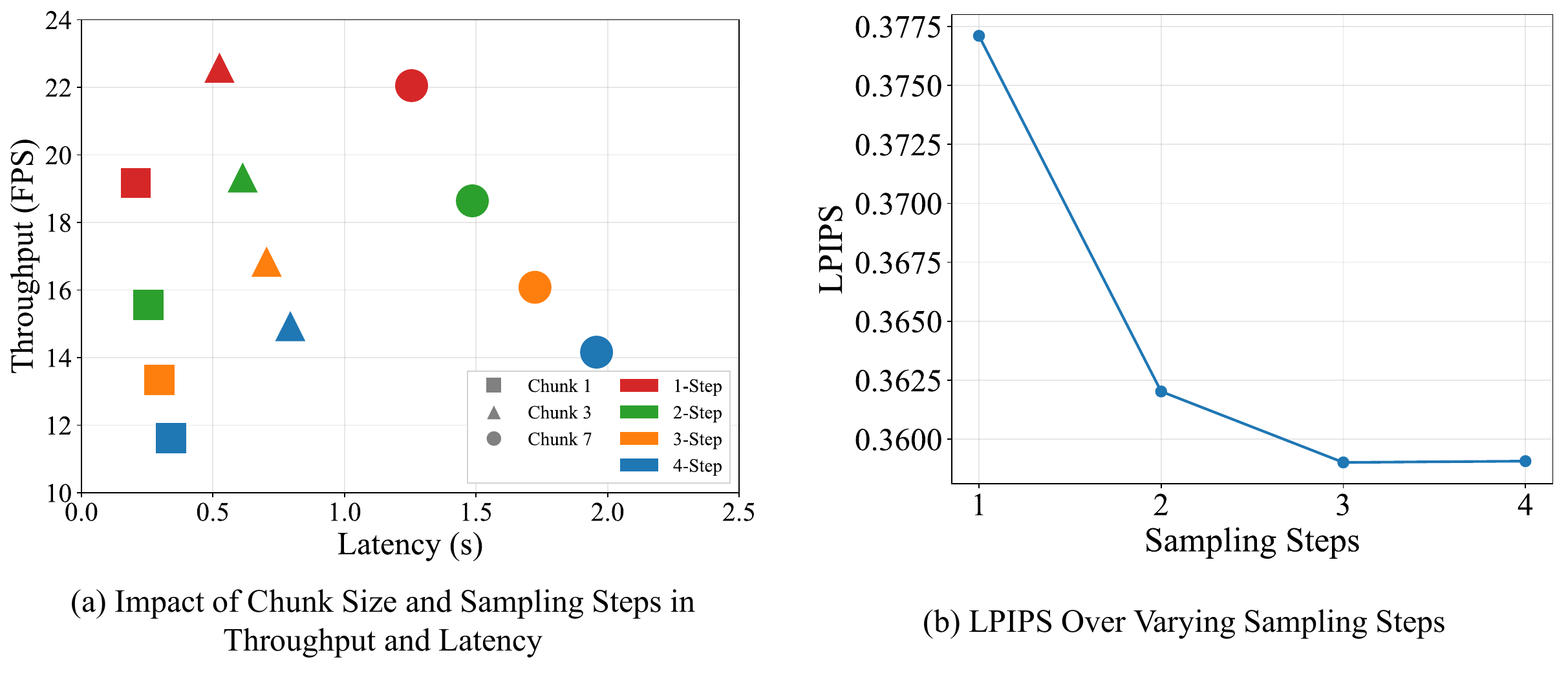}
\caption{\textbf{Speed and Quality Tradeoffs with Chunk Sizes and Sampling Steps.} We visualize the latency-throughput relationship across varying chunk sizes and sampling steps (left), and image quality (LPIPS) across different sampling steps, using our default setup of \texttt{c3s1w1} (right).}
\label{fig:sup_abl_speed_steps}
\vspace{-3mm}
\end{figure}

\noindent\textbf{Impact of Chunk Size and Sampling Steps.} In the main paper, we primarily focused on evaluating the efficiency of diverse sparse attention patterns with attention sink size and attention window size. As shown in Table~\ref{tab:ablation} and Figure~\ref{fig:streaming_abl}, reducing the chunk size increases the number of autoregressive rollouts required to generate the same length of video while reducing the amount of bidirectional attention, leading to worse performance. Since chunk sizes beyond 3 induce high latency, which makes them less useful for our interactive use case, we report only speed metrics for these configurations. While we did not perform controlled training ablations for larger chunk sizes, we observe similar trends with TalkingMachines~\citep{low2025talkingmachines}, where quality generally improves with larger chunks at the cost of latency. Figure~\ref{fig:sup_abl_speed_steps} (a) visualizes the latency and throughput given different chunk sizes and sampling steps. Note that since all benchmarks are conducted with a sink size and a window size of 1, the model with chunk size 7 has lower FPS than the one with chunk size 3 because its attention sequence length is longer.

The number of sampling steps per chunk also affects this balance. We train our student model for 3-step generation, which we find to be optimal for our motion-controlled setup; increasing beyond 3 steps yields only marginal quality gains, while reducing to 2 steps causes a noticeable drop in quality, as shown in Figure~\ref{fig:sup_abl_speed_steps}~(b). Although trained for 3-step inference, the DMD framework allows for flexible sampling configurations at test time. Based on these findings, we use a chunk size of 3 with 3 sampling steps as our primary configuration to achieve a strong balance between interactivity and high-quality video generation.

\begin{figure}[t]
\centering
\includegraphics[width=0.75\textwidth]{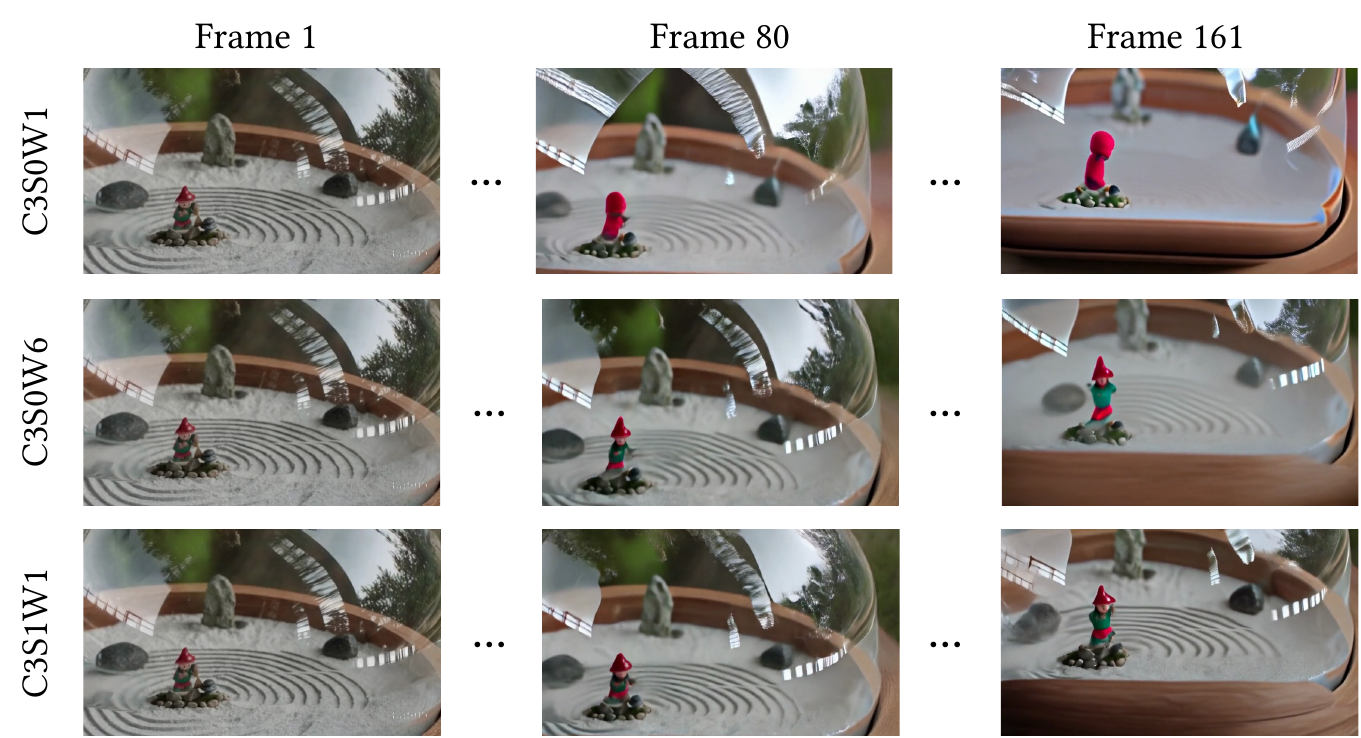}
\caption{\textbf{Long video extrapolation with and without attention sink.} Models without attention sink (top two rows) exhibit cumulative drift over time, while our approach with attention sink (bottom) maintains stable quality throughout extrapolation.}
\label{fig:sup_long_video}
\vspace{-3mm}
\end{figure}

\noindent\textbf{Visualization of Attention Sink's Impact on Long Video Extrapolation.}
We present long video extrapolation results comparing generation with and without attention sinks. As shown in Figure~\ref{fig:sup_long_video}, incorporating at least one sink chunk proves crucial for preventing drift during extended generation. Without this, the model exhibits increasing degradation over time, while the attention sink enables stable quality maintenance throughout the video sequence. Please refer to the videos in the supplementary materials for additional results.

\begin{table}[h]
\scriptsize
\centering
\caption{\textbf{Impact of Motion Control on Generative Quality.} We evaluate whether injecting motion control degrades the pretrained model's capability by comparing against a larger, dedicated I2V baseline (Wan 2.1 14B I2V). We also report performance when motion conditions are dropped. Results indicate that adding motion conditioning does not significantly degrade the base model's generative quality. While removing motion conditions introduces a slight quality drop as our models were not optimized for this setting, the output still adheres to the given text and image inputs.}
\label{tab:i2v_impact_comparison}
\begin{tabular}{lccccccccc}
\toprule
Model & \begin{tabular}[c]{@{}c@{}}i2v\\subject\end{tabular} & \begin{tabular}[c]{@{}c@{}}i2v\\bg\end{tabular} & \begin{tabular}[c]{@{}c@{}}subj.\\consis.\end{tabular} & \begin{tabular}[c]{@{}c@{}}bg\\consis.\end{tabular} & \begin{tabular}[c]{@{}c@{}}motion\\smooth.\end{tabular} & \begin{tabular}[c]{@{}c@{}}aesth.\\qual.\end{tabular} & \begin{tabular}[c]{@{}c@{}}imag.\\qual.\end{tabular} & FVD ($\downarrow$) \\
\midrule
Wan 2.1 I2V (14B) & 0.979 & 0.987 & 0.947 & \textbf{0.953} & 0.988 & 0.619 & \textbf{0.711} & 1274.6 \\
\midrule
Ours Teacher (1.3B) w. Motion & \textbf{0.984} & \textbf{0.988} & \textbf{0.948} & 0.943 & 0.987 & \textbf{0.625} & 0.698 & \textbf{578.2} \\
Ours Teacher (1.3B) w/o Motion & 0.965 & 0.976 & 0.889 & 0.926 & \textbf{0.990} & 0.613 & 0.707 & 1550.4 \\
\midrule
Ours Student (1.3B) w. Motion & 0.982 & 0.987 & 0.940 & 0.941 & 0.985 & 0.618 & 0.684 & 745.8 \\
Ours Student (1.3B) w/o Motion & 0.973 & 0.982 & 0.923 & 0.943 & 0.986 & 0.597 & 0.688 & 1532.5 \\
\bottomrule
\end{tabular}
\end{table}

\noindent\textbf{Impact of Motion Control on Generative Capability.} To assess whether injecting motion control compromises the base model's generative capability, we compare our 1.3B models against the larger Wan 2.1 14B I2V baseline, which serves as a high-quality upper bound (as no Wan 1.3B I2V exists) on the Sora dataset. As shown in Table~\ref{tab:i2v_impact_comparison}, we observe no noticeable degradation; despite the capacity gap and open-source training data, performance remains robust. In fact, a few demo samples demonstrate that motion control enables the model to generate dynamic motions that naive I2V models often struggle to produce due to static bias~\citep{choi2025enhancing}. Consequently, FVD scores are lower with motion conditions, as the generated videos better mimic the target distribution through explicit motion constraints.

We also investigate the model's behavior when motion conditions are not provided. Since our models are incentivized to accurately follow motion and are not explicitly trained for an I2V (or ``motionless'') setup, providing empty motion inputs indeed results in a quality degradation. Nonetheless, while the quality is slightly lower compared to the fully conditioned setting, the model still follows text prompts effectively without severe visual collapse. One interesting observation is that the teacher model without motion conditions rarely produces sudden scene changes, yet results in a lower subject consistency metric. We hypothesize that applying CFG with empty motion conditions leads to unstable outputs where text prompts dominate. We did not observe this behavior with guidance-distilled student models. Please refer to the supplementary videos for detailed results. 

\begin{figure}[t]
\centering
\includegraphics[width=1.0\textwidth]{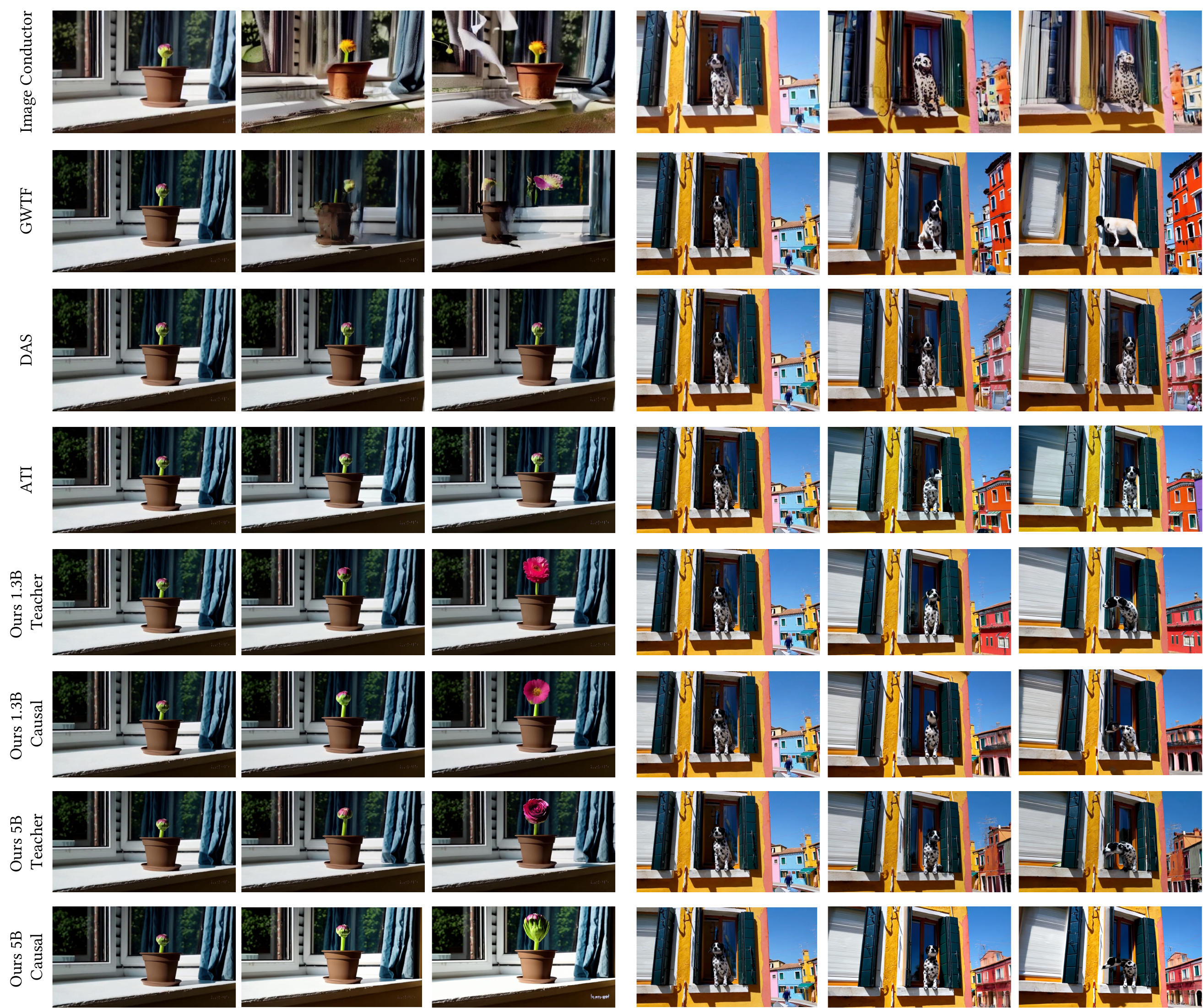}
\caption{\textbf{Qualitative Comparison.} Generated videos from the Sora subset. As seen on the left, our model successfully reconstructs the flower blooming motion, while GWTF captures the motion but suffers quality degradation. ATI produces high-quality videos but with reduced trajectory adherence in motion transfer scenarios.}
\label{fig:sup_qual}
\end{figure}

\noindent\textbf{Qualitative Comparison.} 
We also provide a qualitative comparison between baselines in Figure~\ref{fig:sup_qual}. Please refer to the supplementary videos for detailed results.

\begin{figure}[h]
\centering
\includegraphics[width=1.0\textwidth]{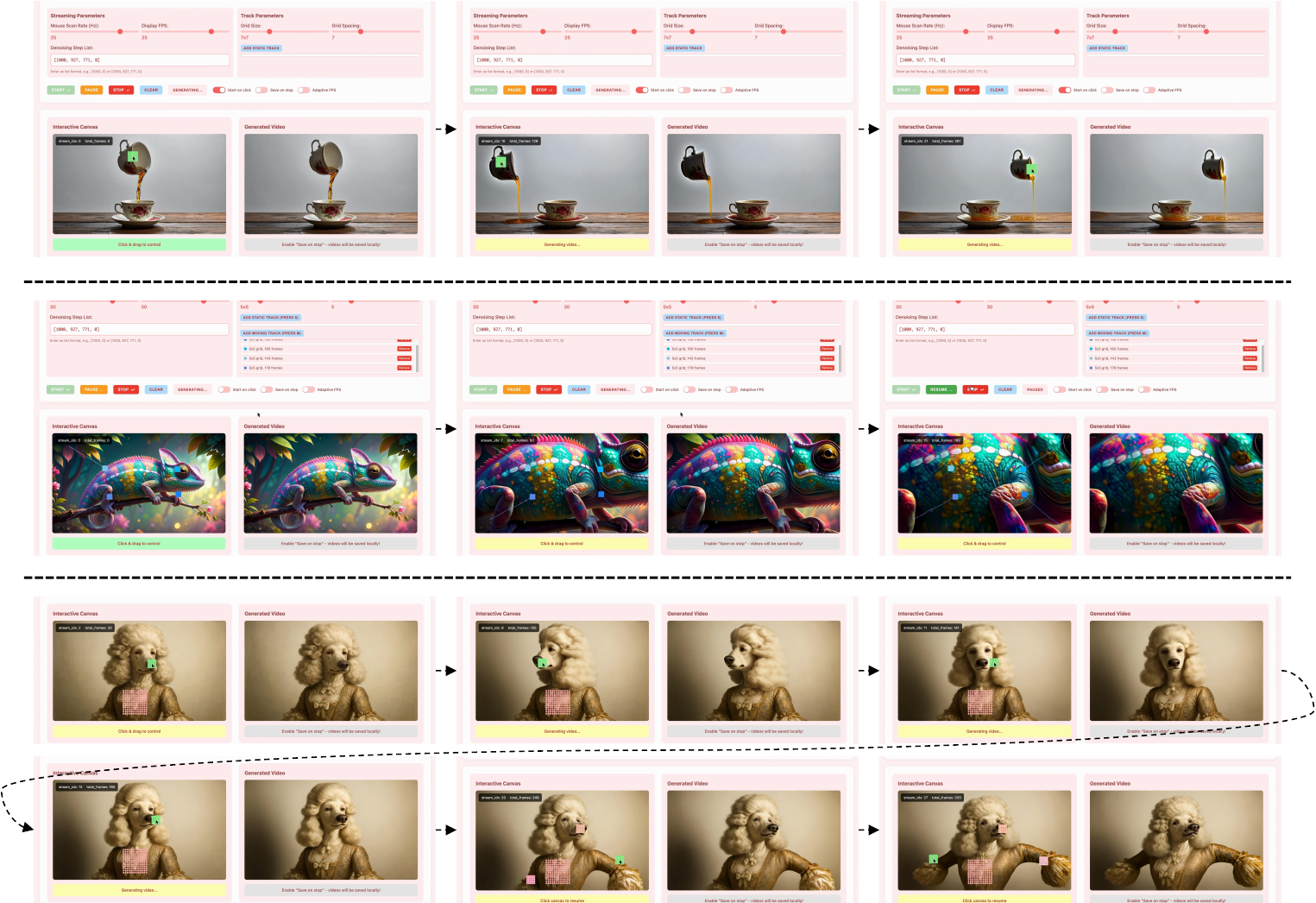}
\caption{\textbf{Streaming Demo.} We show some examples of our streaming demo. Green grids indicate the points that are being dragged (online), red grids indicate a set of points that should remain static, and blue grids indicate user's pre-drawn trajectories for moving multiple points simultaneously (see chameleon example). More examples can be found on our supplementary website.}
\label{fig:sup_streaming_demo}
\end{figure}

\noindent\textbf{Streaming Demo.} We show some examples of our streaming demo in Figure~\ref{fig:sup_streaming_demo}. The demo starts by accepting an input image and text prompts, which can help generate effects that are not achievable through mouse drags. Users can then choose the specific spacing/size of the track grids and start controlling objects in a scene, or move the camera. Due to its autoregressive nature, users can pause or resume the streaming generation process. Users can start/end or pause/resume the generation process with \texttt{Enter} and \texttt{Space} keyboard input, which is especially useful for dynamically adding static grids to specify unmoving regions or multiple moving grids to control different motions during streaming. As a video generative model, our method naturally supports drag-based image editing, generating intermediate transition frames as a bonus, while being faster than most dedicated drag-based image editing methods~\citep{pan2023drag, shi2024dragdiffusion, nie2023blessing, shin2024instantdrag, zhao2024fastdrag}. To further support diverse downstream tasks, we will continuously update our front-end UI with additional features. 

\section{Training Details}
\label{sec:sup_training_details}
\noindent\textbf{Data Preprocessing.}
We train our models on two primary data sources: OpenVid-1M~\citep{nan2024openvid} and synthetic videos generated by larger Wan text-to-video models. For OpenVid-1M, we filter the dataset to 0.6M videos by requiring a minimum of 81 frames and a 16:9 aspect ratio, sampling at 16 FPS. For synthetic data, we use 70K samples for Wan 2.1 (81 frames, 480P resolution, generated by Wan 2.1 14B using text prompts from VidProm~\citep{wang2024vidprom}) and 30K publicly available samples for Wan 2.2 (121 frames, 720P resolution, generated by Wan 2.2 5B from the FastVideo team~\citep{fastvideo2025}).

We extract motion trajectories from all videos using CoTracker3~\citep{karaev2024cotracker3}, tracking points on a 50$\times$50 uniform grid. Our training follows a two-stage design: initial training on OpenVid-1M to establish general motion conditioning capability, followed by fine-tuning on cleaner synthetic data to improve trajectory adherence and reduce artifacts from noisy real-world videos. For the distillation phase, we use only synthetic samples since the distillation process converges quickly and requires less data.

For causal adaptation, we generate 4,000 training samples using our joint guidance strategy $(w_t=3.0, w_m=1.5)$. Notably, Self Forcing-style distillation with DMD objectives does not require complete video sequences during training due to the nature of distribution matching loss design, requiring only the first frame, text prompt, and corresponding motion tracks.

\noindent\textbf{Teacher Model Training.} 
We initialize our teacher models from partial weights of VideoXFun's Wan variants~\citep{videoxfun2024}, which extend Wan I2V models with additional control channels. This initialization accelerates convergence compared to training from scratch. Both Wan 2.1 and Wan 2.2 undergo two-stage training: (1) initial training on filtered OpenVid-1M (0.6M videos) for 4.8K steps, followed by (2) fine-tuning on cleaner synthetic data for 800 steps (Wan 2.1) and 400 steps (Wan 2.2), approximately one epoch each. During training, we randomly sample 1,000$\sim$2,500 tracks and assign a sinusoidal positional embedding of $d=64$ dimensions. Stochastic track masking (described in Sec.~\ref{sec:adding_motion_control}) is applied during the fine-tuning stage. We use batch size 128 with learning rates of $1 \times 10^{-5}$ and $1 \times 10^{-6}$ for stages 1 and 2, respectively. The track head remains frozen after initial training as it already operates chunk-wise.

We train the motion-guided teacher model through rectified flow matching objective, where the forward process linearly interpolates between data $z_0$ and Gaussian noise $z_1 \sim \mathcal{N}(0, I)$: 
\begin{equation}
z_t = (1-t) z_0 + t z_1, \quad t \in [0,1],
\end{equation}
with timestep shifting~\citep{wan2025}, defined as $t’(k,t) = \frac{kt/1000}{1 + (k-1)(t/1000)}$, where $k=6.0$. 
The model is trained to predict the expected velocity fields 
conditioned on noisy latents, denoising time steps, text prompts $c_t$ and motion embedding $c_m$, with timestep-dependent weighting $w_t$:
\begin{equation}
\mathcal{L}_{\text{FM}} = \mathbb{E}_{z_0, z_1, t} \left[ w_t \left\| v_\theta(z_{t’}, t’, c_{t}, c_{m}) - (z_1 - z_0) \right\|^2 \right].
\end{equation}

\paragraph{Causal Architecture Adaptation.} 
Following CausVid~\citep{yin2025causvid}, we adapt to few-step causal inference through ODE trajectory regression. We generate 4,000 ODE trajectories from the teacher model and train for 2,000 steps. While the transition from bidirectional to causal attention requires significant adaptation, we find variations within causal patterns (e.g., different window sizes) are similar enough to learn jointly with single model. We therefore train with diverse sparse causal attention masks, creating a unified initialization that supports flexible self-rollout configurations. We maintain a batch size of 128 with a learning rate $2 \times 10^{-6}$.

\paragraph{Self Forcing-Style Distillation with DMD.} 
Self Forcing distillation converges quickly at around 400 steps with a batch size of 64. We set learning rates to $2 \times 10^{-6}$ for the generator and $4 \times 10^{-7}$ for the critic (fake score function), with a 1:5 update ratio and gradient truncation as described in Sec.~\ref{sec:causal_distillation}.

We employ AdamW optimizer~\citep{kingma2014adam, loshchilov2017decoupled} with mixed precision (bfloat16) and PyTorch's FSDP. Exponential moving average (EMA) is applied during teacher training and DMD distillation. Wan 2.1 variants train at $832 \times 480$ resolution while Wan 2.2 trains at $1280 \times 704$. With 32 A100 GPUs, training the Wan 2.1 teacher model takes roughly 3 days with causal adaptation and distillation completing in 20 hours, while Wan 2.2 requires slightly longer training times.

\section{Evaluation Protocols}
Since different methods employ various backbone models with different spatial and temporal resolutions, we optimize the evaluation setup for each method by matching their primary spatial/temporal dimensions. For DAVIS evaluation, when the number of frames exceeds a model's default temporal length, we retain the first and last frames while uniformly subsampling intermediate frames, then compare against correspondingly subsampled ground truth frames.

All Sora demo videos were limited to 81 frames at 16 FPS for standard experiments, except for the extrapolation experiments in Table~\ref{tab:ablation} and Figure~\ref{fig:streaming_abl}, which use up to 241 frames (average 194 frames, approximately 15 seconds). Similar subsampling was applied for models with shorter temporal contexts (16 frames for AnimateDiff, 49 for CogVideoX).

We report the best scores for each method using its optimal configuration. For Image Conductor, we tested with {1, 10, 100, 1000, 2500} tracks and report results from 100 tracks, which performed best. Go-With-The-Flow requires dense optical flow as input, so we provide flow estimated by RAFT following their reference implementation. Diffusion-As-Shader uses 3D tracks as input, for which we provide tracks from SpatialTracker~\citep{xiao2024spatialtracker} on a $70\times70$ grid (their default setting). ATI was tested with {40, 2500} tracks, and we report results from 40 tracks (their default), which performed better. Our models consistently use 2D trajectories from $50\times50$ initial grid points tracked by CoTracker3.

After generation, all results are resized to $832\times480$ resolution to ensure a consistent scale for metrics, particularly EPE which calculates L2 distance between track coordinates. All latency and throughput are measured using a single H100 GPU in bfloat16 precision with Flash Attention 3~\citep{shah2024flashattention}.

For the camera control benchmark on the LLFF dataset, we derive $50\times50$ grid tracks using the method described in Sec.~\ref{sec:quantitative_eval}. To minimize unintended object motion and focus the evaluation purely on camera movement, we use the prompt template: ``static scene, only camera motion, no object is moving, \texttt{\{scene\_name\}}'', where \texttt{\{scene\_name\}} is replaced with the specific LLFF scene name.

\section{Limitation and Future Work}
\label{sec:sup_limitation}
While MotionStream achieves real-time motion control for long-range video generation, we identify several limitations. First, the fixed attention sink mechanism, while ensuring stable long-term generation, constrains the model's ability to handle scenarios with complete scene changes. Our approach maintains strong anchoring to the initial chunk, which works well for most motion-controlled generation scenarios where cameras and objects move within consistent environments. However, when presented with trajectories from game engines or other sources where environments change continuously, the model exhibits a tendency to preserve the initial scene rather than adapting to new contexts. This limitation is also inherent to current 2D tracking systems, which cannot meaningfully track and encode complete scene transitions. Future work could explore dynamic attention sinking strategies that adaptively refresh anchor frames for world modeling applications.

Second, we observe artifacts when motion trajectories are extremely rapid or physically implausible, manifesting as temporal inconsistencies or distortions in object appearance. One good approach for future work would be exploring effective track augmentation strategies during training to better simulate imperfect user inputs and scaling to larger backbone models, which generally exhibit more robust visual quality. 

\begin{figure}[t]
\centering
\includegraphics[width=1.0\textwidth]{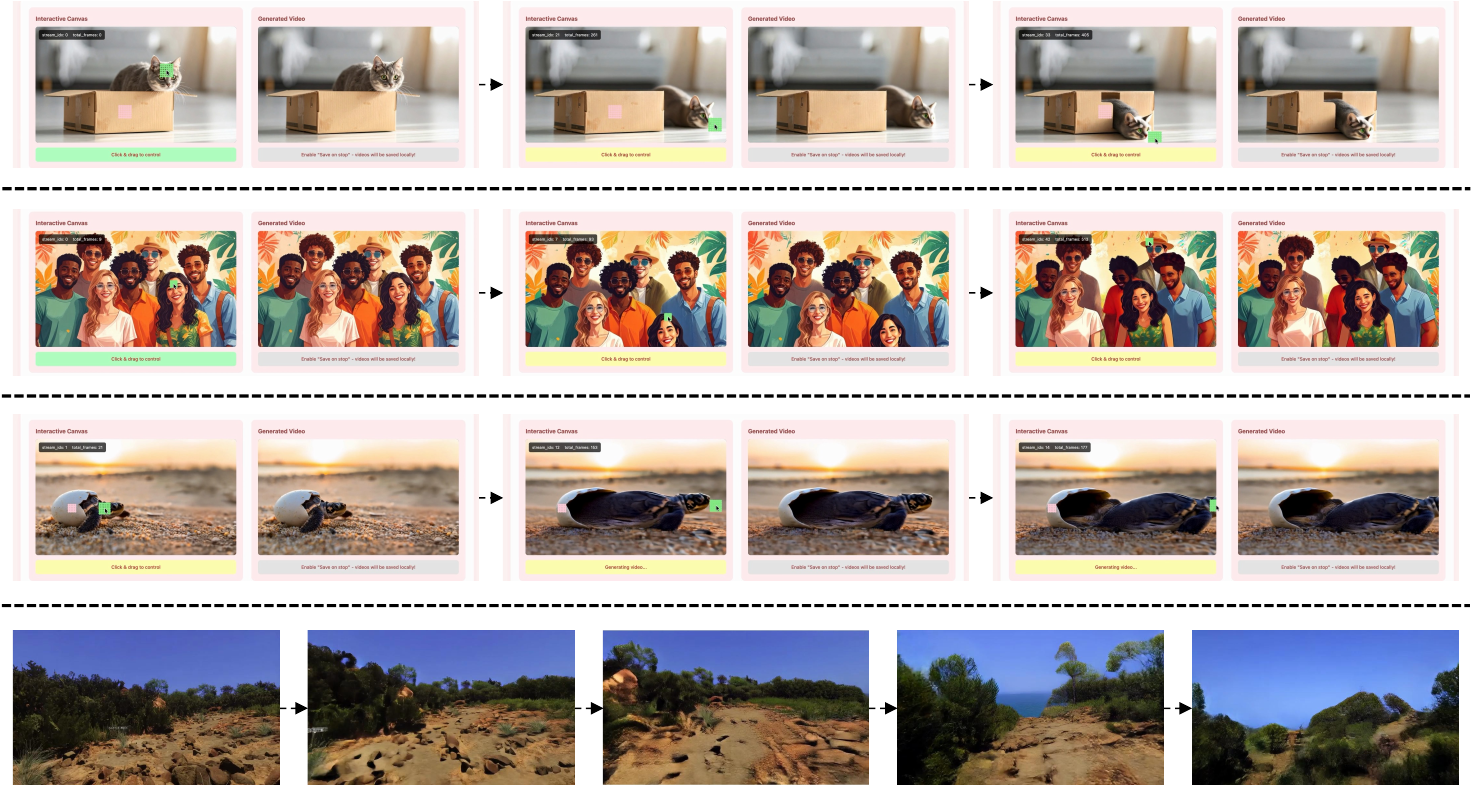}
\caption{\textbf{Failure Cases.} In the cat and turtle examples, the intention was to bring the cat out of the box and make the turtle hatch from the egg. However, due to limitations in hand-drawn trajectories for expressing complex motions and the backbone model's generalization capacity, the outputs are not physically plausible, and objects are either deformed or simply translated along the tracks. In complex scenes with multiple human identities (second example), the model often loses identity information and produces artifacts. The last row shows our model's output on the world exploration task. Since track representation struggles to capture complete scene transitions and the attention sink prioritizes preserving source features, our pipeline faces difficulty in scenarios where new objects appear or scenes continuously change. More videos can be found on our supplementary website.}
\label{fig:sup_failure}
\end{figure}

Lastly, our pipeline sometimes struggles to preserve source details when scenes, text prompts, or intended motions are highly complex. This primarily stems from backbone capacity limitations. While motion conditioning with text prompts can enforce movements beyond what the base model generates from text alone, quality may be unsatisfactory in such cases. We also note that different image conditioning mechanisms across different backbones can affect the robustness of the model in handling imperfect motion cues. Interestingly, we empirically observed that the smaller Wan 2.1 (1.3B) usually outperforms Wan 2.2 (5B) in preserving source structures, particularly with user-drawn flat-grid imperfect point trajectories. We attribute this to Wan 2.1's input image cross-attention design which helps maintain the original structure throughout, while Wan 2.2's TI2V structure is slightly more experimental. As our models are relatively modest in scale, we expect larger base models to provide improved performance and stability under challenging scenarios. Some of our failure cases can be found in Figure~\ref{fig:sup_failure}.

\section{Ethics Statement}
\label{sec:ethics}
As video generative models become increasingly capable of producing realistic content that mimics world dynamics, we recognize the potential for misuse. While \methodname advances interactive content creation with intuitive controls, our approach naturally inherits potential risks from the underlying generative technology, including the creation of deceptive media. We emphasize the critical need for parallel development of safeguarding techniques such as watermarking, content authentication, and controlled access mechanisms. We encourage prioritizing responsible deployment strategies alongside capability improvements to ensure these tools benefit society while minimizing harms.

\noindent\textbf{Reproducibility.} Our models are built upon publicly available Wan model variants and datasets (OpenVid-1M and synthetic Wan videos). Training was conducted in a well-controlled environment, with all training details, hyperparameters, and implementation specifics provided in Sec.~\ref{sec:sup_training_details} and the supplementary to ensure reproducibility.

\noindent\textbf{LLM Usage.} We used LLMs to help polish the writing and presentation of this manuscript. LLMs were not used for research ideation, experimental design, or scientific discovery in this work.

\end{document}